\documentclass[final,5p,times,twocolumn]{elsarticle}

\usepackage{amssymb}

\usepackage{bbm}
\usepackage{amsthm,amsmath,amssymb}
\usepackage{subfigure}
\usepackage{multirow}

\usepackage{threeparttable}
\usepackage{booktabs}

\usepackage[normalem]{ulem}

\journal{arXiv}

\begin{document}

\begin{frontmatter}



\title{Attention-Mamba: A Mamba-Enhanced Multi-Scale Parallel Inference Network for Medical Image Segmentation}


\author[1,2]{Yanhua Zhang}
\ead{yanhuazhang@mail.nwpu.edu.cn}

\author[1]{Ke Zhang\corref{cor1}}
\ead{zhangke@nwpu.edu.cn}

\author[1]{Jingyu Wang}
\ead{jywang@nwpu.edu.cn}

\author[2]{Gabriella Balestra}
\ead{gabriella.balestra@polito.it}

\author[2]{Samanta Rosati}
\ead{samanta.rosati@polito.it}

\author[3]{Yulin Wu\fnref{label1}}
\ead{wuyulin@mail.nwpu.edu.cn}

\author[4]{Wuwei Wang\fnref{label1,label2}}
\ead{wangwuwei124@mail.nwpu.edu.cn}

\author[5,6]{Valentina Giannini\fnref{label2}}
\ead{valentina.giannini@unito.it}

\cortext[cor1]{Corresponding author}

\fntext[label1]{Part of the work was done when they were PhD candidates at Northwestern Polytechnical University.}

\fntext[label2]{W. Wang and V. Giannini contributed equally as last authors.}

\affiliation[1]{organization={School of Astronautics, Northwestern Polytechnical University},
            city={Xi'an},
            postcode={710072},
            country={China}}

\affiliation[2]{organization={Department of Electronics and Telecommunications, Politecnico di Torino},
            city={Turin},
            postcode={10129},
            country={Italy}}

\affiliation[3]{organization={Beijing Aerospace Automatic Control Research Institute},
            city={Beijing},
            postcode={100000},
            country={China}}

\affiliation[4]{organization={Xi'an University of Posts and Telecommunications},
            city={Xi'an},
            postcode={710072},
            country={China}}

\affiliation[5]{organization={Department of Oncology, University of Turin},
            city={Turin},
            postcode={10124},
            country={Italy}}

\affiliation[6]{organization={Candiolo Cancer Institute, FPO-IRCCS},
            city={Candiolo},
            postcode={10060},
            country={Italy}}


\begin{abstract}

\textit{Background and Objective:} U-shaped architectures have long dominated the field of medical image segmentation, while Transformers are widely employed for modeling long-range dependencies. The former typically handles scale variations implicitly by aggregating multi-level features, whereas the efficiency of the latter is constrained by its quadratic computational and memory complexity. 

\textit{Methods:} In this work, we propose an effective alternative to traditional U-shaped architectures by constructing parallel branches at different levels to obtain multi-scale features and corresponding predictions. Furthermore, we enhance our network by integrating Mamba, a state space model that captures long-range dependencies with linear complexity. First, a dual-path architecture with lateral connections aggregates high-level semantic information and low-level spatial details at each branch. Then, we introduce a Recursive Alignment Module (RAM) that restores spatial details in low-resolution features through stepwise alignment, optimizing them for subsequent global feature learning and multi-scale fusion. We further build parallel Mamba branches upon aligned features to establish hierarchical global representations. Finally, we propose a Mamba-based attention mechanism for adaptive multi-scale prediction fusion; this mechanism utilizes Mamba to enhance information exchange across scales along both the channel and spatial dimensions.

\textit{Results:} Compared to state-of-the-art 2D CNN, Transformer, and Mamba-based networks, our model achieves the highest segmentation performance on the Synapse, ACDC, ISIC-2018, and PH2 datasets while maintaining high efficiency, featuring the second-smallest parameters (14.05 M) and moderate computational complexity (8.94 GFLOPs).

\textit{Conclusions:} Experiments across four datasets and three imaging modalities (MRI, CT, and dermoscopy) demonstrate that integrating multi-scale parallel inference with Mamba-based modeling effectively improves segmentation performance while maintaining computational efficiency, indicating good robustness to modality- and task-specific variations. Code is available at: https://github.com/Yanhua-Zhang/Attention-Mamba.

\end{abstract}




\end{frontmatter}


\section{Introduction}
\label{Introduction}
Medical image segmentation is a fundamental task in medical image analysis and a prerequisite for many downstream tasks and clinical practices, including disease quantification, anatomical assessment, and surgical navigation \cite{xing2025segmamba, zheng2025xfmamba, pipoli2025fuse}. Accurate delineation of anatomical structures and pathological regions can reduce clinician workload, improve reproducibility, and mitigate inter- and intra-observer variability \cite{liu2024swin, wang2024mamba}. 

Convolutional neural networks (CNNs), particularly U-Net and its variants \cite{ronneberger2015u,huang2020unet,isensee2021nnu}, have become the dominant paradigm for medical image segmentation due to their ability to capture hierarchical local features. However, their limited receptive field restricts the modeling of long-range dependencies. Transformers address this limitation through self-attention mechanisms \cite{vaswani2017attention}, with some of them still adhering to a U-shaped architecture \cite{liu2025cswin,ren2025hresformer,chen2021transunet}. Moreover, Transformers' quadratic computational and memory complexity with respect to the input size poses challenges for high-resolution medical images or feature maps.

\begin{figure*}[!t]
  \centering
  \subfigure[Image Pyramid]{
  \label{Image-Pyramid}
  \includegraphics[width=2.0in]{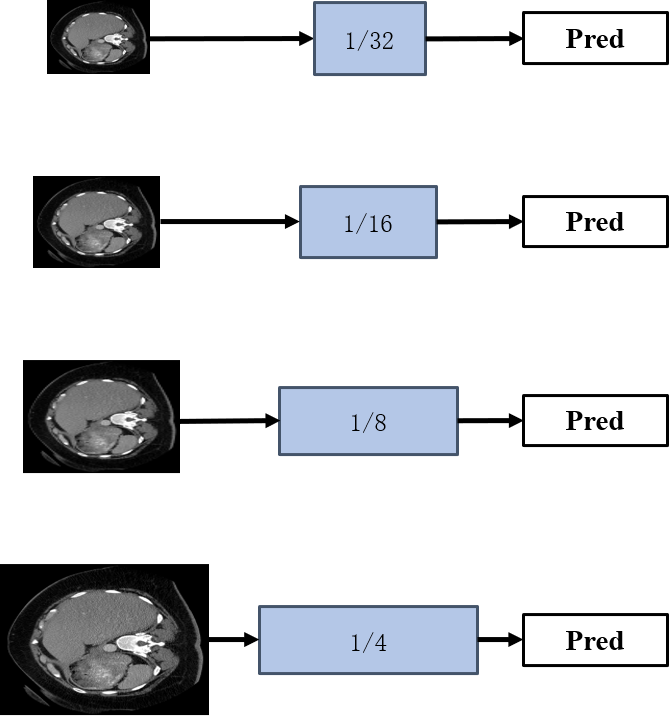}}
  \hfil
  \subfigure[Feature Pyramid Networks]{
  \label{FPN-like}  
  \includegraphics[width=2.0in]{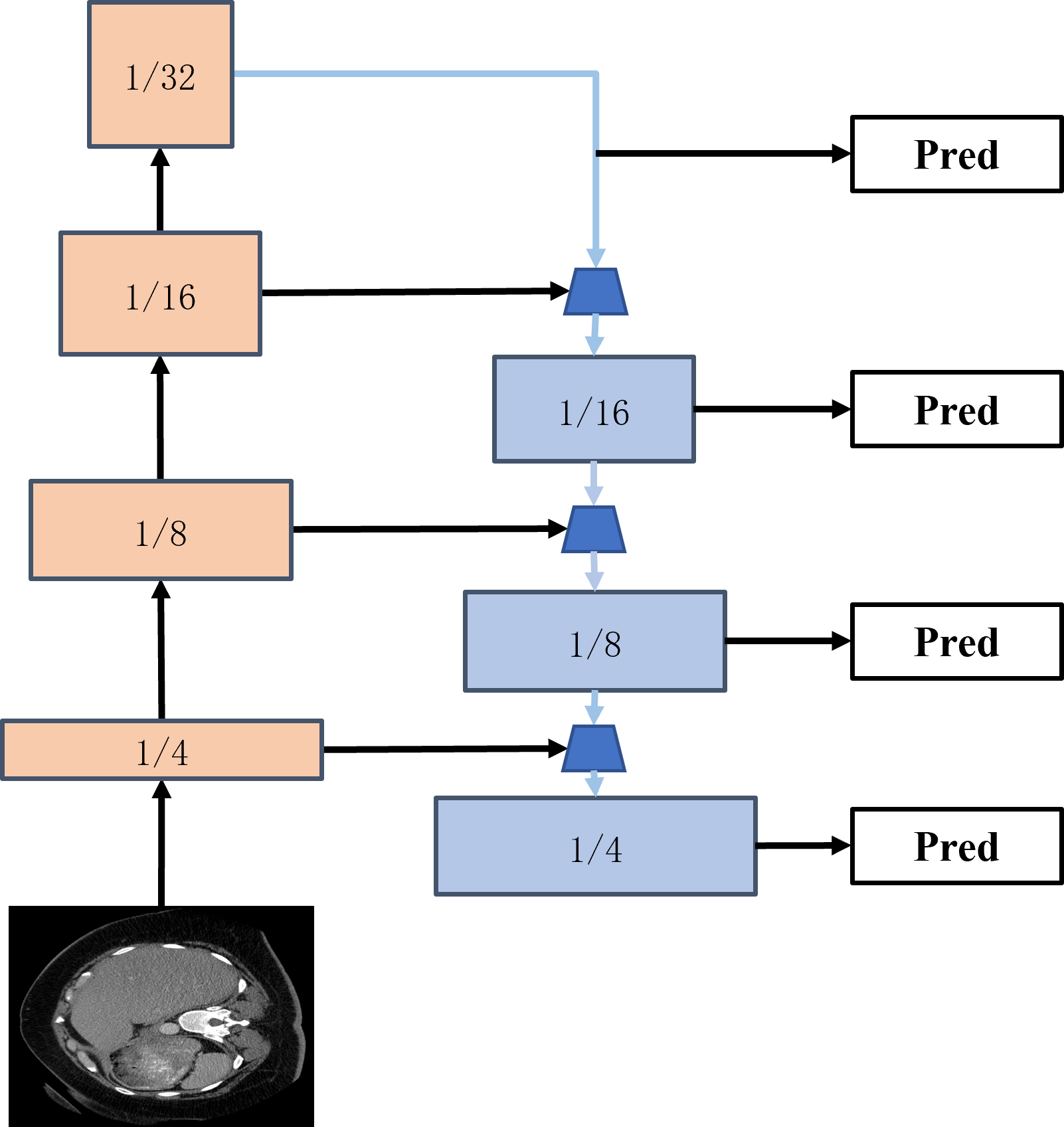}}
  \hfil
  \subfigure[Our Network]{
  \label{Our architecture}
  \includegraphics[width=2.5in]{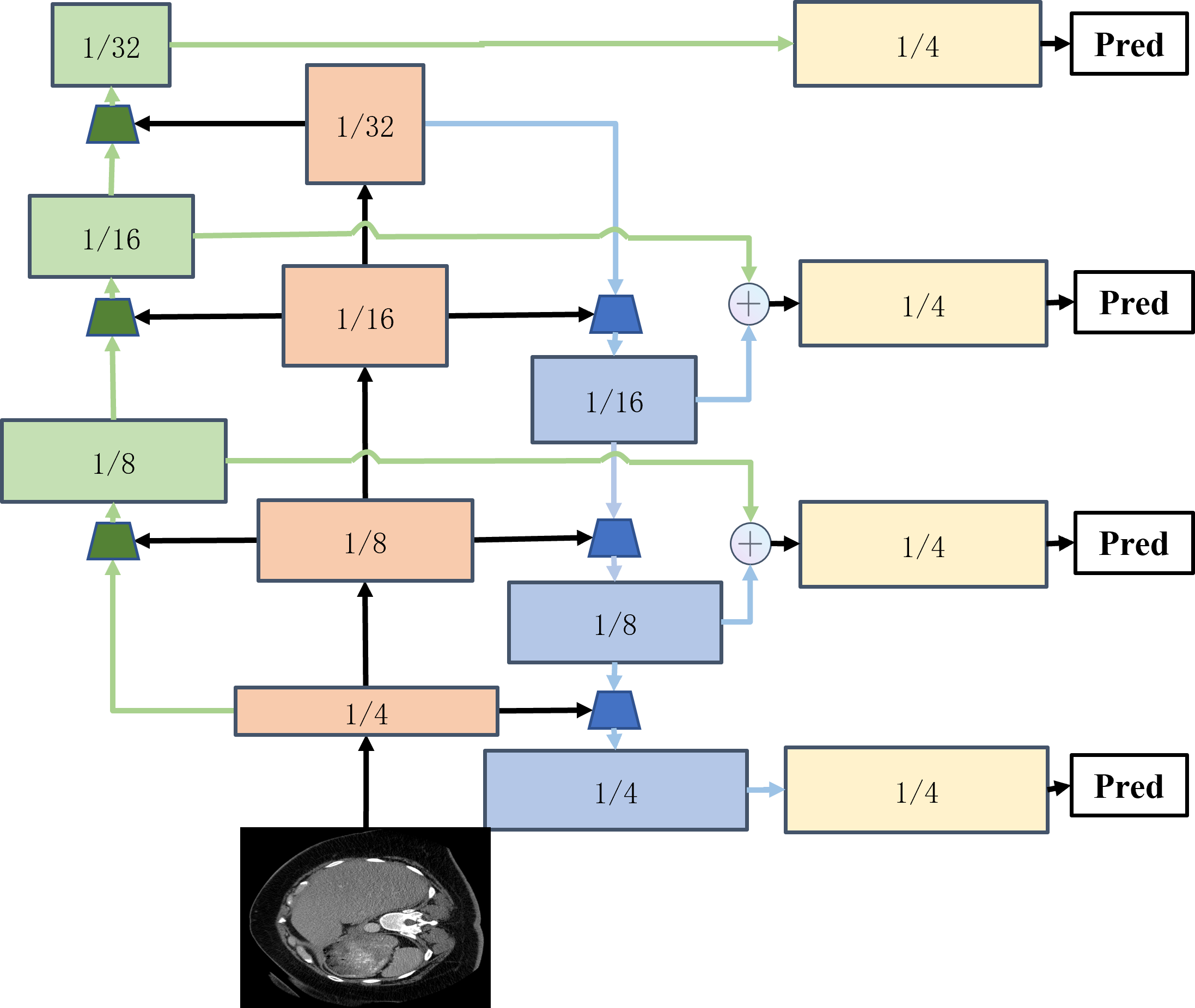}}
  \caption{The comparison between different parallel inference networks.}
  \label{Different parallel inference networks}
  \end{figure*}

A further challenge in medical image segmentation lies in the effective modelling of multi-scale features, which is essential for capturing the significant variability in the size and shape of anatomical structures and lesions. A straightforward approach is the image pyramid strategy, which processes multiple rescaled inputs independently \cite{rahman2024emcad,rahman2023medical,chen2016attention,tao2020hierarchical,fan2020pranet}. While effective in improving performance, this method significantly increases computational complexity and memory overhead. A more efficient alternative is to perform parallel inference at each level of the encoder or decoder to produce multi-scale predictions. This in-network multi-scale inference architecture is widely used in object detection tasks \cite{lin2017feature}, but it has rarely been explored for segmentation tasks. Existing methods, such as PraNet \cite{fan2020pranet}, perform multi-stage predictions and refine them in a coarse-to-fine manner. However, independently generating predictions at each level does not fully exploit the complementary nature of low-level spatial details and high-level semantic information \cite{zhu2025automatic,azad2025transception,jiang2026e2miseg}. To aggregate multi-scale predictions, approaches such as CASCADE \cite{rahman2023medical} employ simple upsampling operations (e.g. bilinear interpolation) followed by summation. These operations are insufficient to address spatial misalignment between features from distant layers, particularly under large upsampling factors. 

More recently, state space models, such as Mamba \cite{gu2024mamba}, have emerged as a strong alternative to Transformers, offering the ability to capture long-range dependencies with linear complexity. Due to its efficiency in processing long sequences, it was soon adapted for 2D vision tasks \cite{wang2024mamba, ma2024u, liu2024vmamba, zhu2024vision, xing2025segmamba}. However, existing approaches predominantly adopt sequential architectures, where Mamba layers are stacked in encoder-decoder frameworks. Additionally, existing Mamba-based feature interaction methods \cite{xu2024polyp, pipoli2025fuse, zheng2025xfmamba} limit flexibility in modeling multi-scale interactions and primarily focus on spatial modeling, often overlooking channel-wise dependencies .  

To address the aforementioned limitations, in this work we propose an innovative unified framework for efficient multi-scale segmentation that consists of four components. First, a Multi-level Feature Aggregation Module (MFAM) integrates hierarchical encoder features at each scale to enhance representation quality. Second, a Recursive Alignment Module (RAM) progressively aligns multi-scale features to reduce spatial inconsistencies. Third, a Mamba-based module is employed to model long-range dependencies through parallel branches at different feature levels, enabling hierarchical global representation learning. Finally, a Mamba-based Cross-scale Attention Module (MCAM) performs adaptive fusion of multi-scale predictions by modeling cross-scale dependencies along both spatial and channel dimensions.

Unlike existing sequential Mamba-based designs, the proposed framework constructs parallel Mamba branches at different levels of the CNN backbone. In addition, RAM transforms high-level, low-resolution features into longer sequences via progressive upsampling while preserving semantic information, thereby alleviating the performance degradation of Mamba layers when operating on short sequences \cite{yu2025mambaout}. Furthermore, MCAM explicitly models cross-scale interactions by compressing features across scales and applying Mamba-based operations along both spatial and channel dimensions, enabling more effective multi-scale prediction fusion.

Our main contributions are summarized as follows.
\begin{enumerate}
\item{The introduction of MFAM to integrate multi-level features through the dual-path architecture and the lateral connections. With this simple and efficient structure, we can aggregate low-level spatial details and high-level semantic information at each scale.}
\item{The use of RAM, designed to perform stepwise alignment using intermediate features, making it more effective and efficient for the alignment between long-distance feature maps. By restoring the spatial details of low-resolution features, it facilitates subsequent global feature learning and multi-scale prediction fusion.}
\item{The replacement of Transformers with Mamba modules to more efficiently model long-range dependencies with linear complexity. Different with the sequential design, we build parallel Mamba branches at different levels of the CNN-part to obtain hierarchical global representations.}
\item{The introduction of a Mamba-based attention mechanism for adaptive multi-scale prediction fusion by generating attention scores based on information from all scales. Within MCAM, the Cross-scale Channel Interaction (CCI) module and the Cross-scale Spatial Interaction (CSI) module employ Mamba to enhance information exchange across scales along the channel and spatial dimensions, respectively.}
\end{enumerate}

\section{Related Works}
\label{Related Works}

\subsection{Mamba-based Networks}
\label{Mamba-based Networks}
Many recent studies in medical image segmentation have focused on designing tailored scanning paths to improve the performance of Mamba-based models. Most of them adopt SS2D to scan the 2D input from four directions as the standard component for building their Mamba-based networks \cite{ruan2024vm, zhang2024vm, liu2024swin}. LKM-UNet \cite{wang2024lkm} employs bidirectional Mamba in its architecture, whereas SegMamba \cite{xing2025segmamba} introduces tri-oriented spatial Mamba for handling 3D inputs. Some works seek to augment Mamba with local feature extraction capabilities. HC-Mamba \cite{xu2024hc} introduces a hybrid block with parallel CNN and Mamba branches to jointly learn local and global representations. Both LoG-VMamba \cite{dang2024log} and HybridMamba \cite{wu2025hybridmamba} utilize LocalMamba \cite{huang2024localmamba} for learning local dependences by restricting the scan path of Mamba in a limited window. Existing Mamba-based networks typically employ a sequential architecture, stacking several Mamba stages successively to form the encoder and reducing the feature resolution after each stage. Differently, we build parallel  Mamba branches upon upsampled features for high-resolution hierarchical global feature learning.

Mamba can also be used to facilitate cross-scale feature interactions. Both Polyp-Mamba \cite{xu2024polyp} and SMM-UNet \cite{li2025selective} employ a similar design: they concatenate upsampled or downsampled features from different encoder levels into a sequence, and then feed this sequence into Mamba layers to enhance the interaction between features at different scales. To enhance the interaction among very long sequences, IM-Fuse \cite{pipoli2025fuse} propose an interleaved concatenation strategy by arranging tokens from different modality-features alternately before feeding them into Mamba layers. Inspired by \cite{wan2025sigma}, XFMamba \cite{zheng2025xfmamba} leverages the system matrix C of State Space Models (SSMs) as an agent to facilitate information exchange among features from different branches. The methods mentioned above only perform cross-scale information exchange along the spatial dimension, while our proposed MCAM model enables feature interaction along both the channel and spatial dimensions.

\subsection{Parallel/Multi-scale Inference}
\label{Parallel/Multi-scale Inference}
The image pyramid (Fig. \ref{Image-Pyramid}) offers a direct solution for handling scale variability in computer vision \cite{chen2016attention, zhao2017pyramid, tao2020hierarchical}. Chen et al. \cite{chen2016attention} introduced an attention mechanism for the adaptive fusion of score maps, whereas Tao et al. \cite{tao2020hierarchical} leveraged a sequential chain structure to handle multi-scale outputs. For medical image segmentation, FCT \cite{tran2021tmd}, TMD-Unet \cite{tragakis2023fully}, and Canet \cite{xie2023canet} augment the encoder with a pyramid image input to enhance robustness against variations in object scale. Furthermore, the multi-scale image pyramid is frequently adopted as a training strategy in polyp segmentation frameworks \cite{fan2020pranet,rahman2023medical,rahman2024emcad}. To promote efficiency, some networks perform parallel inference on different levels of the encoder to produce several score maps \cite{hariharan2015hypercolumns,yang2020small,ding2020semantic,ding2022sab}. NDNet \cite{yang2020small} employs the convolutional operation to achieve weighted integration of multi-scale predictions, while CGBNet \cite{ding2020semantic} and SABNet \cite{ding2022sab} adopt higher-resolution intermediate-level predictions to progressively refine the coarse, low-resolution prediction from the top level. In medical imaging, both PraNet \cite{fan2020pranet} and CASCADE \cite{rahman2023medical} perform inference at each level of the encoder or decoder to obtain multi-scale predictions. The former refines predictions progressively via a coarse-to-fine strategy, whereas the latter directly fuses multi-resolution predictions after bilinear upsampling. 

This in-network parallel inference architecture has rarely been explored in the field of medical imaging. In this work, we enhance this type of architecture from three aspects: multi-level feature aggregation, spatial alignment, and adaptive multi-scale prediction fusion.

\begin{figure*}[!t]
  \centering
  \includegraphics[width=0.97\linewidth]{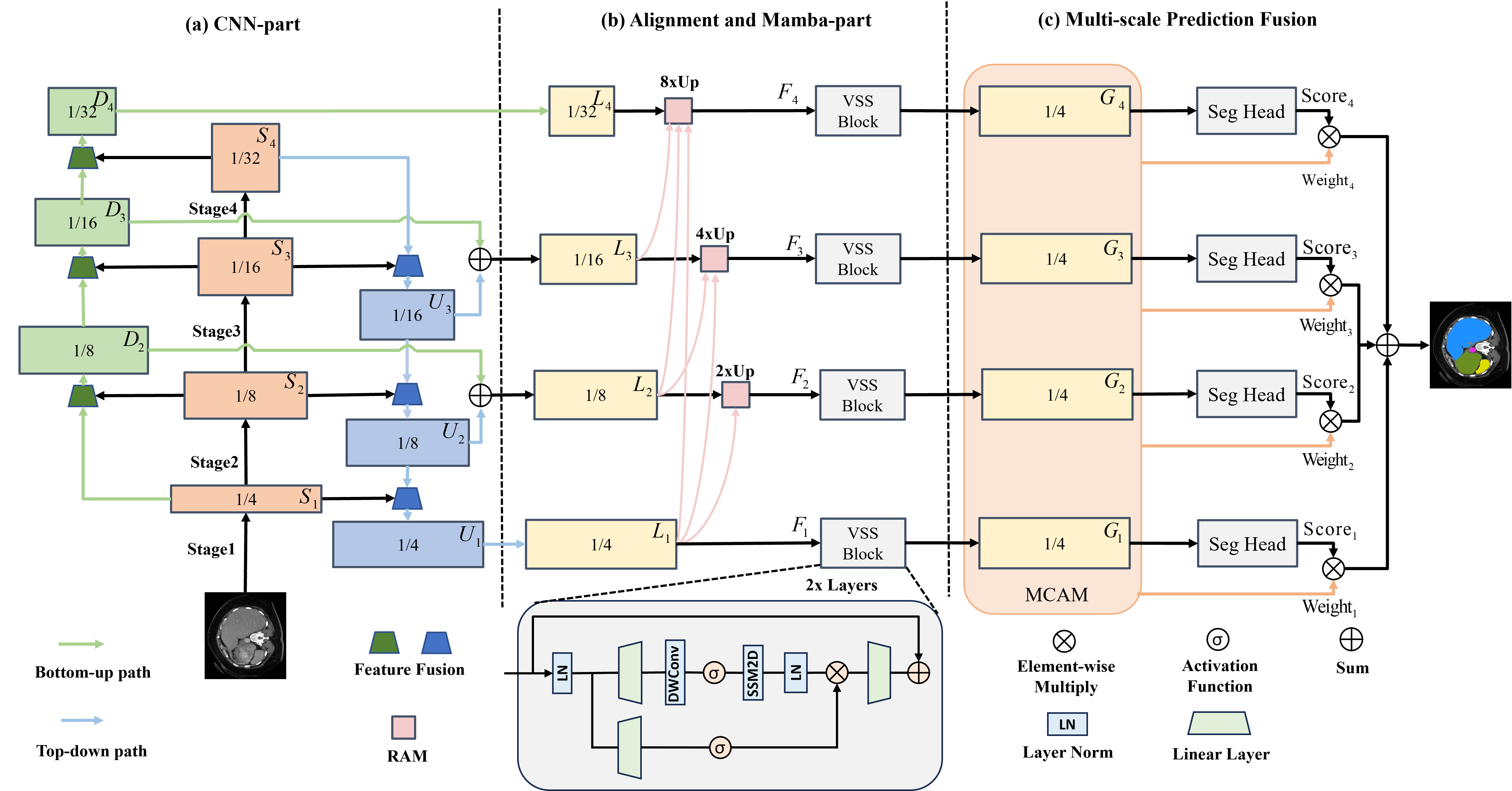}
  \caption{Overall architecture of our approach. The Recursive Alignment Module (RAM) and the Mamba-based Cross-scale Attention Module (MCAM) are illustrated in Fig. \ref{Recursive Alignment} and Fig. \ref{MCAM}, respectively.}
  \label{Overall}
\end{figure*}

\subsection{Feature Alignment/Upsampling}
\label{Feature Alignment/Upsampling}
Bilinear interpolation is a common choice for upsampling feature or score maps due to its simplicity and efficiency \cite{ronneberger2015u, zhou2018unet++, zhang2024multitrans}. However, its data-independent nature makes the precise alignment of features across different resolutions challenging, particularly in the presence of large displacements. In order to restore the spatial details of low-resolution feature maps, Noh et al. \cite{noh2015learning} introduced the deconvolution layer, which was widely adopted by U-Net and its followers in their decoder \cite{ronneberger2015u,isensee2021nnu}. To ensure efficiency, SegNet \cite{badrinarayanan2017segnet} reuses pooling indices from the encoder to perform nonlinear upsampling of decoder features. GUN \cite{mazzini2018guided} employs a transformation algorithm to learn 2D spatial shifts for adaptive feature alignment. Inspired by the optical flow methods \cite{brox2004high, zhu2017deep}, both AlignSeg \cite{huang2021alignseg} and FaPN \cite{huang2021fapn} incorporate alignment modules that treat feature warping as a flow-estimation task, allowing for data-driven interpolation. All methods mentioned above are used to align adjacent features with small resolution differences. For medical imaging, the optical flow has also influenced various learning-based methods for image registration \cite{siebert2022learn, balakrishnan2019voxelmorph}, motion estimation \cite{ye2023sequencemorph, joshi2025interpolai}, and disease progression assessment \cite{liu2025imageflownet, wang2023strainnet}. To handle large displacements between images, they are typically based on a complete architecture with both an encoder and a decoder, which are too heavy to integrate into segmentation networks for feature-level alignment.  

The proposed RAM is a lightweight module that leverages intermediate features to handle large displacement alignment, allowing it to be seamlessly integrated into existing network architectures.

\section{Methods}
\label{Our Network}

\subsection{Overview}
\label{Overview}

Figure \ref{Overall} presents the overall framework of the proposed CNN–Mamba hybrid model. The CNN-part is mainly used for local-feature extraction and multi-level feature aggregation. It uses top-down and bottom-up paths to aggregate hierarchical encoder features at each scale, thereby providing each branch with both high-level semantic context and low-level spatial detail. Then, we designed a flow-based recursive alignment module to address the spatial misalignment between different branches. Upon the aligned features, we placed Mamba layers at different levels to provide global representations for the prediction of each branch. Finally, we designed a Mamba-based cross-scale attention module to adaptively fuse these predictions.

\subsection{Preliminaries}
\label{Preliminaries}

To provide a theoretical foundation for the proposed network, we first review the principle of Structured State Space Sequence Models (S4) \cite{gu2021efficiently} and Mamba \cite{gu2024mamba}. These models are inspired by the classical continuous system, in which a one-dimensional input $x(t)\in\mathbb{R}$ is transformed into an output $y(t) \in \mathbb{R}$ via intermediate hidden states $h(t) \in {\mathbb{R}^N}$. This process is represented as a linear ordinary differential equation (ODE):
\begin{equation}
\label{ODE}
\begin{array}{*{20}{c}}
{h'(t) = {\rm{A}}h(t) + {\rm{B}}x(t)}\\
{y(t) = {\rm{C}}h(t)}
\end{array}
\end{equation}
here, $A \in {\mathbb{R}^{N \times N}}$ is the state matrix, while $B \in {\mathbb{R}^{N \times 1}}$ and $C \in {\mathbb{R}^{N \times 1}}$ represent projection matrices. 

To make the continuous system suitable for deep learning applications, S4 and Mamba perform discretization by the zero-order hold (ZOH) method. By introducing a timescale parameter $\Delta $, matrices $A$ and $B$ are converted into their discrete counterparts as follows: 
\begin{equation}
\label{discrete counterparts}
\begin{array}{*{20}{c}}
{\bar A = \exp (\Delta A)}\\
{\bar B = {{(\Delta A)}^{ - 1}}(\exp (\Delta A) - I) \cdot \Delta B}
\end{array}
\end{equation}

Then, Eq. \ref{ODE} can be rewritten as:
\begin{equation}
\label{discrete eq}
\begin{array}{*{20}{c}}
{h(t) = {\rm{\bar A}}h(t - 1) + {\rm{\bar B}}x(t)}\\
{y(t) = {\rm{C}}h(t)}
\end{array}
\end{equation}

Furthermore, the iterative process can be computed using global convolutions to improve computational efficiency, as defined below:
\begin{equation}
\label{global convolution}
\begin{array}{*{20}{c}}
{\bar K = (C\bar B,C\overline {AB} ,...,C{{\bar A}^{L - 1}}\bar B)}\\
{y = x * \bar K}
\end{array}
\end{equation}
where $\bar K \in {\mathbb{R}^L}$ represents a structured convolutional kernel, $ * $ refers to convolution operation, and $L$ is the length of the input. 

Readers are recommended to refer to \cite{gu2021efficiently, gu2024mamba} for more details.

\subsection{Networks}
\label{Networks}

\subsubsection{Multi-level Feature Aggregation Module (MFAM)}
\label{Efficient Multi-level Feature Aggregation module(MFA)}

We employ ResNet-18 as the encoder to extract hierarchical feature representations, which has four sequential stages (${S_1} \sim {S_4}$) with output spatial resolutions of 1/4, 1/8, 1/16, and 1/32 of the original input. Before being processed by MFAM, features from encoder are passed through a $1 \times 1$ convolutional layer for channel adjustment, which is omitted for simplicity.

Upon encoder features, we utilize a top-down path to aggregate mid-level features and pass them to the lowest-level feature. In this process, we use bilinear interpolation to upsample low-resolution features, followed by an element-wise summation and a $3 \times 3$ convolutional layer for fusion:
\begin{equation}
\label{Feature Fusion1}
{U_n} = {\rm{Con}}{{\rm{v}}_{3 \times 3}}({\rm{Up}}({U_{n + 1}}) + {\rm{Con}}{{\rm{v}}_{1 \times 1}}({S_n})),n = 3,2,1.
\end{equation}
Here, ${U_4} = {\rm{Con}}{{\rm{v}}_{1 \times 1}}({S_4})$. Likewise, we build a bottom-up path to integrate low-level features to the highest-level feature. Differently, we utilize bilinear interpolation to downsample features instead of upsampling:
\begin{equation}
\label{Feature Fusion2}
{D_{n+1}} = {\rm{Con}}{{\rm{v}}_{3 \times 3}}({\rm{Down}}({D_{n}}) + {\rm{Con}}{{\rm{v}}_{1 \times 1}}({S_{n+1}})),n = 1,2,3{\rm{,}}
\end{equation}
where ${D_1} = {\rm{Con}}{{\rm{v}}_{1 \times 1}}({S_1})$. 

The intermediate features in the top-down and bottom-up paths still lack features from certain levels (e.g., ${D_2}$ fuses ${S_1}$, ${S_2}$ but lacks ${S_3}$, ${S_4}$). To address this, we introduce two lateral connections between the two paths to facilitate information exchange, enabling the mid-level features (${F_2}$ and ${F_3}$) to incorporate information from all hierarchical levels:
\begin{equation}
\label{lateral connections}
{F_n} = {\rm{Con}}{{\rm{v}}_{3 \times 3}}({D_n} + {U_n}),n = 2,3.
\end{equation}

\begin{figure}[!t]
  \centering
  \subfigure[]{
  \label{Straightforward Alignment}
  \includegraphics[width=0.96\linewidth]{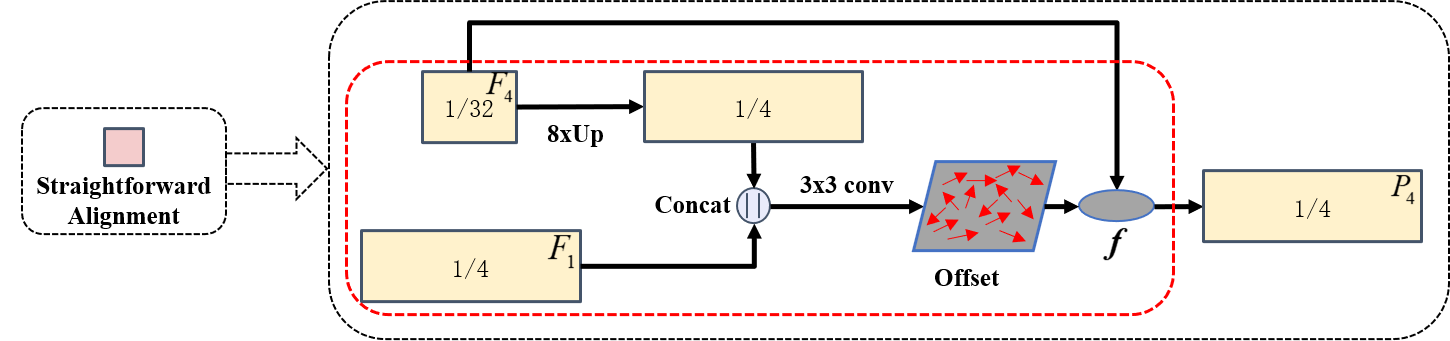}}
  \hfil
  \subfigure[]{
  \label{Recursive Alignment}
  \includegraphics[width=0.98\linewidth]{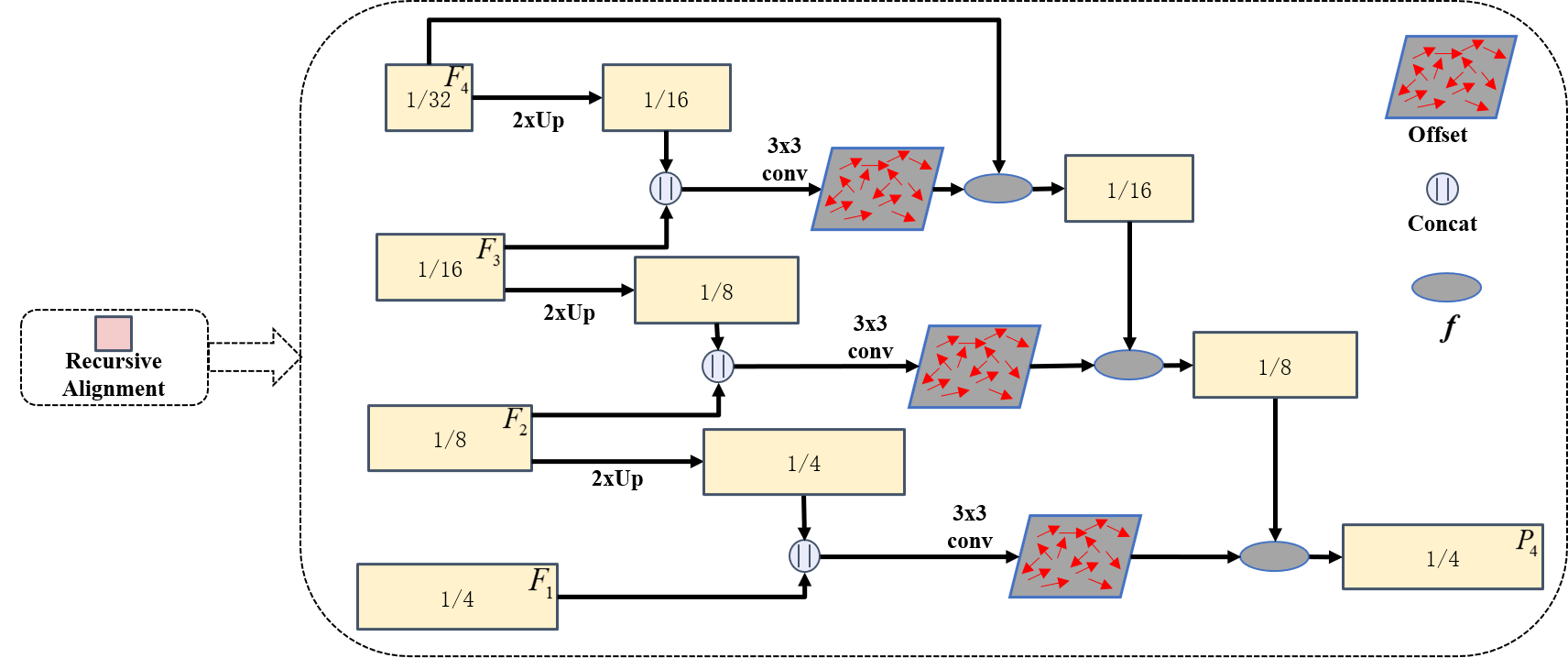}}
  \caption{The comparison between straightforward alignment (top) and recursive alignment (bottom). Here, $f$ refers to the differentiable bi-linear sampling function defined in Eq. \ref{differentiable bi-linear sampling mechanism}.}
  \label{The comparison between Straightforward Alignment and Recursive Alignment}
  \end{figure}

\subsubsection{Recursive Alignment Module (RAM)}
\label{Recursive Alignment Module (RAM)}

The proposed RAM is used to guide the alignment between low-resolution features and the highest-resolution feature, especially for large displacements (e.g., between ${F_4}$ and ${F_1}$). As illustrated in Fig. \ref{Recursive Alignment}, we adopt intermediate features to achieve recursive alignment. A 3 $\times$ 3 convolutional layer is first applied to compute the offset $\Delta$ between the high-resolution feature ${F_n}$ and low-resolution feature ${F_{n + 1}}$:
\begin{equation}
  \label{rewritten}
  {\Delta _n} = {\rm{Con}}{{\rm{v}}_{3 \times 3}}({\rm{Concat}}({F_n},{\rm{Up}}({F_{n + 1}}))),n = 1,2,3.
  \end{equation}
Here, ${\rm{Concat}}$ denotes channel-wise concatenation, while ${\rm{Up}}$ indicates bilinear upsampling. Then, the warped grid between ${F_{n + 1}}$ and ${F_n}$ is computed as follows:
\begin{equation}
\label{warped grid}
{\rm{war}}{{\rm{p}}_n} = \frac{{{g_n} + {\Delta _n}({g_n})}}{2},n = 1,2,3,
\end{equation}
where ${g_n}$ represents each position in spatial grid; ${\Delta _n}({g_n})$ is the normalized offset. After obtaining the warped grids between ${F_1}$ and ${F_2}$, ${F_2}$ and ${F_3}$, ${F_3}$ and ${F_4}$, we align ${F_4}$ and ${F_3}$ to ${F_1}$ in a step-wise manner, respectively. It is worth noting that these warped grids only need to be calculated once and can be shared by all alignment processes. The above procedures can be formulated as follows:
\begin{equation}
  \label{an example of aligning}
  {\hat F_n} = {\left\{ {{{\hat f}_i} = f({{\hat f}_{i + 1}},{\rm{war}}{{\rm{p}}_i}),i = n - 1, \cdots ,1} \right\}_{n = 3,4}},
  \end{equation}
where ${\hat f_i}$ represents the intermediate aligned features, with the initial condition ${\hat f_n} = {F_n}$; ${\hat F_n}={{\hat f}_1}$ denotes the final aligned feature map of ${F_n}$. The notation $\left\{  \cdot  \right\}$ indicates the recursive process, and $f$ refers to the alignment function. We employ the differentiable bilinear sampling function \cite{jaderberg2015spatial} to perform alignment:
\begin{equation}
\label{differentiable bi-linear sampling mechanism}
{\hat f_n} = f({f_n},{\rm{war}}{{\rm{p}}_n}) = \sum\limits_{i \in {\rm{\cal N}}({\rm{war}}{{\rm{p}}_n})} {{w_i}{f_n}(i)} ,
\end{equation}
where ${\rm{\cal N}}({\rm{war}}{{\rm{p}}_n})$ indicates the neighbors of ${\rm{war}}{{\rm{p}}_n}$ in ${f_n}$, and ${w_i}$ is the kernel weight estimated by warped grid.

\begin{figure*}[!t]
  \centering
  \includegraphics[width=0.99\linewidth]{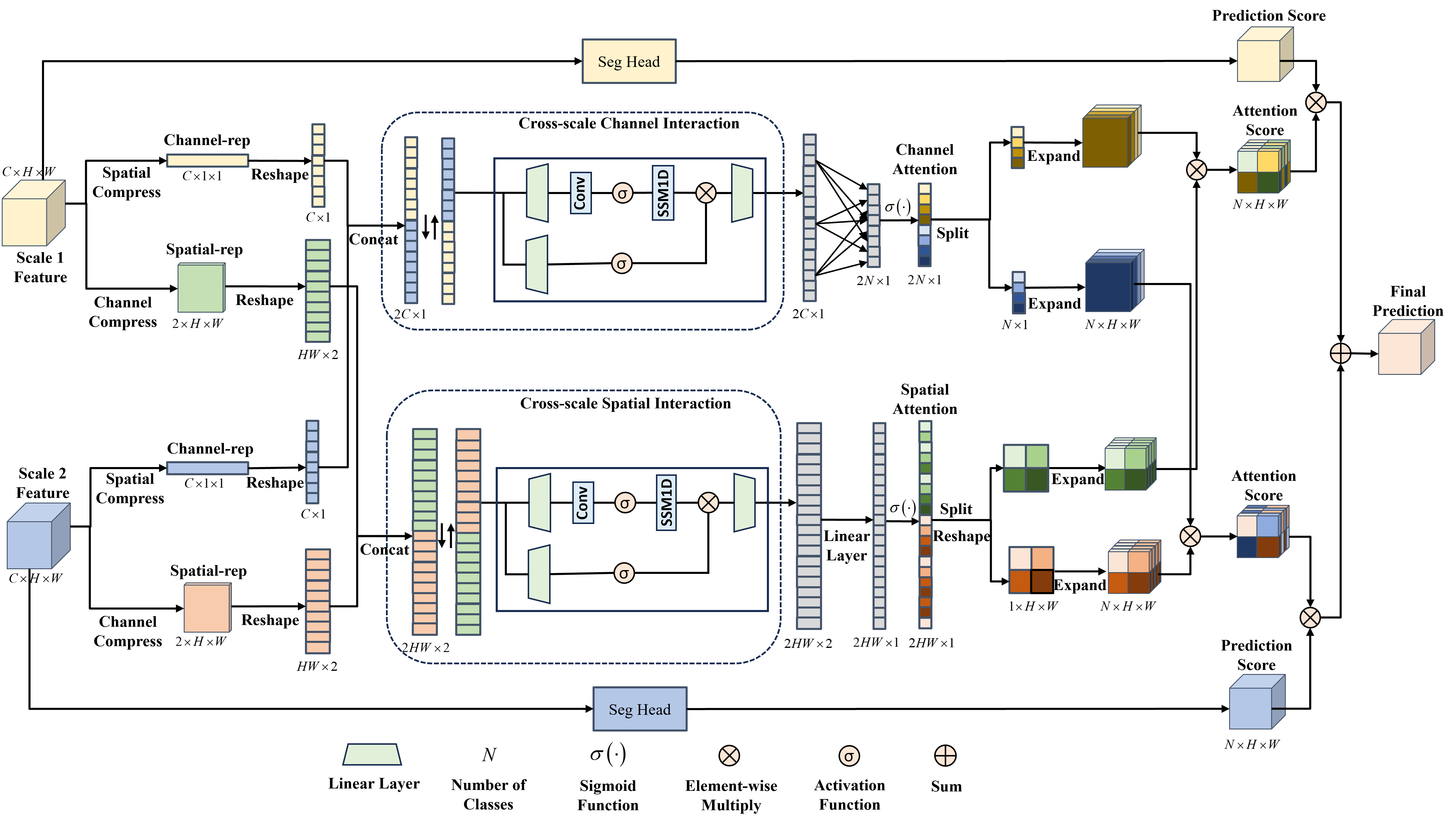}
  \caption{Mamba-based Cross-scale Attention Module.}
  \label{MCAM}
\end{figure*}

Here, we give computational complexity analysis. Since the complexity of other operations in the alignment module is negligible compared to the 3 $\times$ 3 convolutional operation, the overall complexity can be estimated by 3 $\times$ 3 convolutional operations. Therefore, the complexity of computing the offset between ${F_4}$ and ${F_3}$ can be expressed as $W \times H \times {C_{in}} \times {C_{out}} \times 3 \times 3$, which can be simplified as $9WH{C^2}$. Here, $H$ and $W$ denote the height and width of ${F_3}$, and ${C_{in}}$ and ${C_{out}}$ represent the channels of input and output features of the convolutional layer. Since the height and width of ${F_1}$ are increased by 4 times compared to ${F_3}$, the complexity of inputting upsampled ${F_4}$ and ${F_1}$ into the 3 $\times$ 3 convolutional layer can be expressed as $4W \times 4H \times {C_{in}} \times {C_{out}} \times 3 \times 3$. Ultimately, the ratio between the overall complexity of RAM and the straightforward alignment method is: $(9WH{C^2} + 4 \times 9WH{C^2} + 16 \times 9WH{C^2})/(3 \times 16 \times 9WH{C^2}) = 7/16$.

\subsubsection{Multi-level Parallel Mamba Branches }
\label{Multi-level Parallel Mamba Branches }

We introduce a Mamba-based parallel design that efficiently learns multi-scale global representations by operating on features extracted from different levels of the CNN backbone. To further mitigate Mamba's limitations in short sequences \cite{yu2025mambaout}, we construct four parallel Mamba branches on aligned features, enabling effective long-sequence modeling with linear complexity, as follows:
\begin{equation}
  \label{Parallel Mamba Branches}
  \begin{array}{*{20}{c}}
    {G_l} = {\rm{VS}}{{\rm{S}}_{ \times 2}}({\hat F_l}),l = 1,2,3,4.
    \end{array}
  \end{equation}
Here, we use a variation the original Mamba, VSS \cite{liu2024vmamba}, to learn long-range dependences. The core design of VSS is the scan-expansion operation, which performs feature extraction along four scanning paths over the 2D inputs: from top-left to bottom-right, bottom-right to top-left, top-right to bottom-left, and bottom-left to top-right. This design is more effective for processing visual data than the original Mamba, as it enables each pixel to interact with its surroundings in all spatial directions. Given a 2D input feature $s$, the core component of VSS (SS2D) can be expressed as:
\begin{equation}
  \label{SS2D}
\begin{array}{*{20}{c}}
{{s_1},{s_2},{s_3},{s_4} = \rm{expand}(s)}\\
{{{\tilde s}_i} = S6({s_i}),i = 1,2,3,4.}\\
{\tilde s = {\rm{merge}}({{\tilde s}_1},{{\tilde s}_2},{{\tilde s}_3},{{\tilde s}_4})}
\end{array}
\end{equation}
where ${s_1},{s_2},{s_3},{s_4}$ represent four different scanning paths, while ${\rm{merge}}( \cdot )$ includes reshape and sum operations. S6 is the core operation of the original Mamba. Other designs of the VSS block is detailed in Fig. \ref{Overall}.

\subsubsection{Adaptive Scores Fusion Module}
\label{Attention-based Adaptive Multi-scores Fusion}

Predictions at different scales capture complementary information. High-resolution features preserve fine spatial details, while low-resolution features provide richer contextual information. Therefore, we employed an adaptive fusion module to integrate multi-scale predictions. Motivated by \cite{tao2020hierarchical,chen2016attention}, we adopt a pixel-wise attention strategy for this adaptive fusion process. Differently, we proposed a novel Mamba-based Cross-scale Attention Module (MCAM) for generating attention scores for prediction from different branches. Specifically, the Cross-scale Channel Interaction (CCI) module and the Cross-scale Spatial Interaction (CSI) module employ Mamba to enhance information exchange across features of different scales along the channel and spatial dimensions, respectively. This cross-scale information exchange is essential, as it enables the module to generate attention scores for a given scale based not only on its own features, but also on information from other scales. 

Fig. \ref{MCAM} illustrates the adaptive fusion of predictions from two scales, which we extend to support the fusion of four scales. The features from different scales are first compressed along the spatial dimensions independently to generate their respective channel representations, as follows:
\begin{equation}
  \label{channel representations}
  \begin{array}{*{20}{c}}
    R_c^l = {\rm{Av}}{{\rm{g}}_s}({G^l}) + {\rm{Ma}}{{\rm{x}}_s}({G^l}),l = 1,2,3,4,
    \end{array}
  \end{equation}
Here, ${G^l} \in {\mathbb{R}^{C \times H \times W}}$ denotes the feature from the $l$-th scale, and $R_c^l \in {\mathbb{R}^{C \times 1 \times 1}}$ represents its channel representation. ${\rm{Av}}{{\rm{g}}_s}( \cdot )$ and ${\rm{Ma}}{{\rm{x}}_s}( \cdot )$ indicate spatial average pooling and spatial max pooling, respectively.

Similarly, the features are compressed along the channel dimension to obtain their spatial representations:
\begin{equation}
  \label{spatial representations}
  \begin{array}{*{20}{c}}
  R_s^l = {\rm{Concat}}[{\rm{Av}}{{\rm{g}}_c}({G^l}),{\rm{Ma}}{{\rm{x}}_c}({G^l})],l = 1,2,3,4,
    \end{array}
  \end{equation}
where $R_s^l \in {\mathbb{R}^{2 \times H \times W}}$, and ${\rm{Concat}}$ refers to concatenation along the channel dimension. The channel representations from all scales are then reshaped and concatenated into a 1-D sequence, respectively, which are subsequently fed into a Mamba block for information exchange. The original Mamba is designed for 1D sequences with a single direction, which learns an asymmetric, causal relationship \cite{yu2025mambaout}. As relationships between features from different scales are symmetrical,we reversed the concatenated channel and spatial representations to facilitate bidirectional interaction:
\begin{equation}
  \label{channel information exchange}
  \begin{array}{*{20}{c}}
  {{{\mathord{\buildrel{\lower3pt\hbox{$\scriptscriptstyle\leftarrow$}} 
  \over R} }_c} = {\rm{Concat}}[R_c^1,R_c^2,R_c^3,R_c^4]}\\
  {{{\tilde R}_c} = {\rm{S6}}({{\mathord{\buildrel{\lower3pt\hbox{$\scriptscriptstyle\leftarrow$}} 
  \over R} }_c}) + {\rm{S6}}({{\vec R}_c})}
  \end{array}
\end{equation}
where ${\tilde R_c} \in {\mathbb{R}^{4C \times 2}}$. The same operation is applied to the spatial representations: 
\begin{equation}
  \label{spatial information exchange}
  \begin{array}{*{20}{c}}
  {{{\mathord{\buildrel{\lower3pt\hbox{$\scriptscriptstyle\leftarrow$}} 
  \over R} }_s} = {\rm{Concat}}[R_s^1,R_s^2,R_s^3,R_s^4]}\\
  {{{\tilde R}_s} = {\rm{S6}}({{\mathord{\buildrel{\lower3pt\hbox{$\scriptscriptstyle\leftarrow$}} 
  \over R} }_s}) + {\rm{S6}}({{\vec R}_s})}
  \end{array}
\end{equation}
where ${\tilde R_s} \in {\mathbb{R}^{4HW \times 2}}$. Next, the interacted channel representation is passed through two fully connected layers to reduce the channel dimension to the number of classes, and is then processed by a sigmoid function to generate the channel attention scores. Similarly, the channel dimension of the interacted spatial representation is compressed by a linear transformation and then passed through a sigmoid activation to generate the spatial attention scores. These processes are as follows: 
\begin{equation}
\label{attention scores}
\begin{array}{*{20}{c}}
{{A_c} = \sigma {\rm{(FCN(}}{{\tilde R}_c}))}\\
{{A_s} = \sigma {\rm{(}}{{\rm{W}}_s}{{\tilde R}_s})}
\end{array}
\end{equation}
Here, ${A_s} \in {^{4HW \times 1}}$, ${A_c} \in {\mathbb{R}^{4N \times 1}}$, and $N$ is the number of classes. ${\rm{FCN}}$ refers to the fully connected layers. ${{\rm{W}}_s} \in {\mathbb{R}^{2 \times 1}}$ is the weight of the linear layer. Furthermore, the spatial and channel attention scores are redistributed to the four scales: $A_c^1,A_c^2,A_c^3,A_c^4$ and $A_s^1,A_s^2,A_s^3,A_s^4$. Then, the channel and spatial scores of the same scale are expanded and multiplied to generate the attention map for that scale. At the same time, a 1 $\times$ 1 convolutional layer is used as the segmentation head to obtain the prediction score for this scale.
\begin{equation}
  \label{scores}
\begin{array}{*{20}{c}}
{{A^l} = {\rm{Ex}}{{\rm{p}}_s}(A_c^l) \odot {\rm{Ex}}{{\rm{p}}_c}(A_s^l),l = 1,2,3,4,}\\
{{P^l} = {\rm{Con}}{{\rm{v}}_{1 \times 1}}({G^l}),l = 1,2,3,4,}
\end{array}
\end{equation}
where $ \odot $ is element-wise multiplication. ${\rm{Ex}}{{\rm{p}}_s}$ and ${\rm{Ex}}{{\rm{p}}_c}$ denote the expand operations along the spatial and channel dimensions, respectively.
Finally, we use the attention maps to perform a weighted fusion of the multi-scale scores to generate the final prediction:
\begin{equation}
\label{linearly fuse multi-scale scores}
{P_{{\rm{Final}}}} = \sum\limits_{l = 1}^4 {{P_l} \odot {A_l}} .
\end{equation}

\subsection{Multi-scale Supervision}
\label{Multi-scale Supervision}

Consistent with the designed multi-scale prediction architecture, we apply supervision at each individual scale as well as on the fused prediction. In addition, auxiliary supervision is added to each branch after the alignment module to improve gradient flow and offer anatomical guidance for learning alignment offsets. The overall supervisions can be written as:  
\begin{equation}
  \label{overall supervisions}
  {{\cal L}_{{\rm{Total}}}}{\rm{ = }}{\cal L}\left( {G,{P_{{\rm{Final}}}}} \right) + \sum\limits_{l = 1}^4 {{\cal L}\left( {G,{P^l}} \right)}  + \sum\limits_{l = 1}^4 {\lambda _{{\rm{Aux}}}^l{\cal L}\left( {G,P_{{\rm{Aux}}}^l} \right)} ,
  \end{equation}
where $G$ means ground-truth, and $P_{{\rm{Aux}}}^l$ denotes the auxilairy prediction upon ${F_l}$ at the $l$-th scale. $\lambda _{{\rm{Aux}}}^l$ are the weights of auxiliary losses, which are empirically set to 0.25. ${\cal L}$ refers to the objective function used for a specific task. 

For multi-organ segmentation and cardiac structure segmentation tasks, we train our network using a combination of Dice loss and cross-entropy (CE) loss, as follows:
\begin{equation}
  \label{objective function}
  L = {\phi _{{\rm{CE}}}}(G,P) + {\phi _{{\rm{DICE}}}}(G,P),
  \end{equation}
As for skin lesion segmentation, following \cite{ruan2022malunet}, we adopt Dice loss and binary cross-entropy (BCE) loss:
\begin{equation}
  \label{objective function 2}
  L = {\phi _{{\rm{DICE}}}}(G,P) + {\phi _{{\rm{BCE}}}}(G,P).
  \end{equation}

\begin{table*}[!t]\small
  \centering
  \caption{Comparison with the state-of-art methods on the Synapse dataset. The best and second-best values are bolded and underlined, respectively.}
  \label{Table_Synapse}
  \renewcommand\arraystretch{1.0}
  \setlength{\tabcolsep}{0.9mm}{
\begin{tabular}{lllllllllllllll} 
\hline
\multirow{2}{*}{Name} & \multirow{2}{*}{\begin{tabular}[c]{@{}l@{}}DSC \\(\%)\end{tabular}} & \multirow{2}{*}{\begin{tabular}[c]{@{}l@{}}IoU \\(\%)\end{tabular}} & \multirow{2}{*}{\begin{tabular}[c]{@{}l@{}}HD95 \\(mm)\end{tabular}} & \multicolumn{8}{l}{DSC fo each organ} & \multirow{2}{*}{\begin{tabular}[c]{@{}l@{}}Param \\(M)\end{tabular}} & \multirow{2}{*}{\begin{tabular}[c]{@{}l@{}}FLOPs \\(G)\end{tabular}} & \multirow{2}{*}{\begin{tabular}[c]{@{}l@{}}Training\\Size\end{tabular}} \\ 
\cline{5-12}
 &  &  &  & Sp & Ki(R) & Ki(L) & Ga & Li & St & Ao & Pa &  &  &  \\ 
\hline
\multicolumn{15}{l}{CNN-based:} \\ 
\hline
Unet \cite{ronneberger2015u} & 71.45 & 61.45 & 66.5 & 76.1 & 64.77 & 69.15 & 53.15 & 88.89 & 70.92 & 89.49 & 59.16 & 31.04 & 41.88 & $224 \times 224$ \\
AttenUNet \cite{schlemper2019attention} & 77.89 & 68.44 & 60.18 & 83.99 & 75.81 & 76.44 & 62.86 & 94.95 & 73.52 & 90.87 & 64.65 & 34.88 & 50.98 & $224 \times 224$ \\
UNetPlusPlus \cite{zhou2018unet++} & 77.42 & 67.42 & 57.94 & 84.25 & 73.11 & 77.19 & 63.8 & 94.71 & 74.09 & 90.09 & 62.15 & \textbf{9.16} & 26.71 & $224 \times 224$ \\
DualAttenUNet \cite{fu2019dual} & 77.12 & 67.97 & 49.25 & 83.03 & 75.19 & 76.13 & 62.35 & 93.42 & 73.61 & 91.07 & 62.13 & 19.57 & 22.10 & $224 \times 224$ \\ 
\hline
\multicolumn{15}{l}{Transformer-based:} \\ 
\hline
MISSFormer \cite{huang2022missformer} & 76.45 & 66.42 & 60.07 & 81.3 & 75.48 & 80.73 & 60.9 & 92.3 & 72.14 & 88.69 & 60.09 & 42.46 & 7.28 & $224 \times 224$ \\
SwinUNet \cite{cao2022swin} & 78.03 & 67.62 & 52.93 & 83.41 & 69.73 & 78.07 & 62.45 & 92.79 & 77.62 & 89.71 & 70.48 & 27.17 & 5.95 & $224 \times 224$ \\
H2Former \cite{he2023h2former} & 78.99 & 70.05 & 44.89 & 80.03 & 80.56 & 82.34 & 59.63 & 92.76 & 81.89 & 90.36 & 64.33 & 33.68 & 24.70 & $224 \times 224$ \\
HiFormer \cite{heidari2023hiformer} & 83.01 & 74.59 & 32.63 & 90.19 & 83.59 & 85.62 & \textbf{70.23} & 95.85 & 77.12 & 91.23 & 70.28 & 37.46 & 17.51 & $224 \times 224$ \\
MulitTrans \cite{zhang2024multitrans} & 84.71 & 76.66 & 35.21 & 87.63 & \textbf{85.94} & \textbf{87.66} & \uline{65.86} & 94.75 & 83.47 & \textbf{92.2} & \textbf{80.16} & 39.38 & 18.28 & $224 \times 224$ \\ 
\hline
\multicolumn{15}{l}{Mamba-based:} \\ 
\hline
Mamba\_UNet \cite{wang2024mamba} & 81.42 & 72.38 & 39.05 & 86.76 & 78.19 & 80.67 & 65.64 & 95.67 & 82.7 & 90.49 & 71.25 & 19.12 & 3.56 & $224 \times 224$ \\
VM\_UNet \cite{ruan2024vm} & 79.42 & 70.55 & 39.45 & 84.43 & 78.73 & 78.44 & 61.38 & 94.13 & 77.88 & 88.46 & 71.93 & 27.43 & \textbf{3.16} & $224 \times 224$ \\
VM\_UNet\_V2 \cite{zhang2024vm} & 81.4 & 72.43 & 35.37 & 87.76 & 81 & 85.75 & 64.99 & 95.65 & 77.21 & 89.89 & 68.96 & 22.77 & \uline{3.41} & $224 \times 224$ \\
Swin\_UMamba \cite{liu2024swin} & \uline{84.78} & \uline{76.71} & \uline{29.97} & \uline{91.08} & 82.65 & 86.71 & \uline{65.86} & \uline{95.94} & \textbf{86.18} & \uline{92.11} & 77.74 & 59.89 & 33.57 & $224 \times 224$ \\
Our & \textbf{85.62} & \textbf{77.84} & \textbf{24.22} & \textbf{91.21} & \uline{85.63} & \uline{87.31} & \textbf{70.23} & \textbf{96.27} & \uline{84.1} & 91.97 & \uline{78.26} & \uline{14.05} & 6.87 & $224 \times 224$ \\
\hline
\end{tabular}
  }
  \end{table*}


\begin{table*}[!t]\small
  \centering
  \caption{Comparison with the state-of-art methods on the ACDC dataset. The best and second-best values are bolded and underlined, respectively.}
  \label{Table_ACDC}
  \renewcommand\arraystretch{1.0}
  \begin{threeparttable}
  \setlength{\tabcolsep}{0.9mm}{
\begin{tabular}{lllllllllllll} 
\hline
\multirow{3}{*}{Name} & \multicolumn{3}{l}{Val} & \multicolumn{6}{l}{Test} & \multirow{3}{*}{\begin{tabular}[c]{@{}l@{}}Param \\(M)\end{tabular}} & \multirow{3}{*}{\begin{tabular}[c]{@{}l@{}}FLOPs \\(G)\end{tabular}} & \multirow{3}{*}{\begin{tabular}[c]{@{}l@{}}Training\\ Size\end{tabular}} \\ 
\cline{2-10}
 & \multirow{2}{*}{\begin{tabular}[c]{@{}l@{}}DSC\\~(\%)\end{tabular}} & \multirow{2}{*}{\begin{tabular}[c]{@{}l@{}}IoU \\(\%)\end{tabular}} & \multirow{2}{*}{\begin{tabular}[c]{@{}l@{}}HD95 \\(mm)\end{tabular}} & \multirow{2}{*}{\begin{tabular}[c]{@{}l@{}}DSC \\(\%)\end{tabular}} & \multirow{2}{*}{\begin{tabular}[c]{@{}l@{}}IoU \\(\%)\end{tabular}} & \multirow{2}{*}{\begin{tabular}[c]{@{}l@{}}HD95 \\(mm)\end{tabular}} & \multicolumn{3}{l}{DSC of Cardiac Structures} &  &  &  \\ 
\cline{8-10}
 &  &  &  &  &  &  & RV & ~MYO & ~LV &  &  &  \\ 
\hline
\multicolumn{13}{l}{CNN-based:} \\ 
\hline
Unet \cite{ronneberger2015u} & 90.95 & 83.86 & 5.93 & 90.82 & 83.84 & 4.7 & 87.9 & 88.95 & 95.6 & 31.04 & 30.75 & $192 \times 192$ \\
AttenUNet \cite{schlemper2019attention} & 91.56 & 84.93 & 5.53 & 91.36 & 84.73 & 5.45 & 88.5 & 89.69 & 95.89 & 34.88 & 37.45 & $192 \times 192$ \\
UNetPlusPlus \cite{zhou2018unet++} & 91.48 & 84.83 & 5.85 & 91.74 & 85.27 & 8.85 & 89.52 & \uline{89.78} & 95.91 & \textbf{9.16} & 19.62 & $192 \times 192$ \\
DualAttenUNet \cite{fu2019dual} & \uline{91.81} & 85.24 & \textbf{4.13} & \uline{91.85} & \uline{85.37} & 8.37 & \uline{89.92} & 89.69 & 95.94 & 19.57 & 16.23 & $192 \times 192$ \\ 
\hline
\multicolumn{13}{l}{Transformer-based:} \\ 
\hline
MISSFormer \cite{huang2022missformer} & 88.51 & 79.99 & 8.16 & 88.23 & 79.71 & 15.52 & 84.33 & 85.74 & 94.61 & 42.46 & 7.28 & $224 \times 224$ \\
SwinUNet \cite{cao2022swin} & 89.74 & 81.93 & 5.66 & 89.7 & 81.95 & 11.37 & 87.29 & 87.51 & 94.3 & 27.17 & 5.95 & $224 \times 224$ \\
H2Former \cite{he2023h2former} & 90.99 & 84.03 & 4.69 & 91.55 & 84.92 & 7.94 & 89.52 & 89.19 & \uline{95.96} & 33.68 & 18.12 & $192 \times 192$ \\
HiFormer \cite{heidari2023hiformer} & 91.23 & 84.3 & \uline{4.54} & 91.24 & 84.46 & \uline{4.27} & 89 & 89.04 & 95.7 & 37.46 & 17.51 & $224 \times 224$ \\
MulitTrans \cite{zhang2024multitrans} & \uline{91.81} & \uline{85.33} & 5.37 & 91.76 & 85.3 & 5.57 & 89.71 & 89.57 & \textbf{95.99} & 39.38 & 13.43 & $192 \times 192$ \\ 
\hline
\multicolumn{13}{l}{Mamba-based:} \\ 
\hline
Mamba\_UNet \cite{wang2024mamba} & 89.63 & 81.79 & 6.5 & 89.77 & 82.19 & 6.44 & 86.21 & 87.72 & 95.39 & 19.12 & 2.60 & $192 \times 192$ \\
VM\_UNet \cite{ruan2024vm} & 90.32 & 82.88 & 5.37 & 90.38 & 83 & 5.57 & 87.88 & 88.16 & 95.11 & 27.43 & \textbf{2.32} & $192 \times 192$ \\
VM\_UNet\_V2 \cite{zhang2024vm} & 91.71 & 85.14 & 4.76 & 91.28 & 84.37 & \textbf{3.47} & 89.71 & 88.78 & 95.34 & 22.77 & \uline{2.48} & $192 \times 192$ \\
Swin\_UMamba \cite{liu2024swin} & 91.61 & 84.98 & 6.39 & 91.65 & 85.07 & 4.4 & 89.74 & 89.36 & 95.86 & 59.89 & 24.65 & $192 \times 192$ \\
Our & \textbf{91.94} & \textbf{85.61} & 4.61 & \textbf{92.03} & \textbf{85.68} & 5.52 & \textbf{90.19} & \textbf{89.91} & \textbf{95.99} & \uline{14.05} & 5.04 & $192 \times 192$ \\
\hline
\end{tabular}
  }
  \end{threeparttable}
  \end{table*}

\begin{table*}[!t]\small
  \centering
  \caption{Comparison with the state-of-art methods on the ISIC-2018 and PH2 dataset. The best and second-best values are bolded and underlined, respectively.}
  \label{Table_ISIC-2018}
  \renewcommand\arraystretch{1.0}
  \begin{threeparttable}
  \setlength{\tabcolsep}{0.9mm}{
\begin{tabular}{llllllllllllllll} 
\hline
Dataset & \multicolumn{6}{l}{ISIC2018} & \multicolumn{6}{l}{PH2} & \multirow{2}{*}{\begin{tabular}[c]{@{}l@{}}Param \\(M)\end{tabular}} & \multirow{2}{*}{\begin{tabular}[c]{@{}l@{}}FLOPs \\(G)\end{tabular}} & \multirow{2}{*}{\begin{tabular}[c]{@{}l@{}}Training \\Size\end{tabular}} \\ 
\cline{1-13}
Method & \begin{tabular}[c]{@{}l@{}}DSC \\(\%)\end{tabular} & \begin{tabular}[c]{@{}l@{}}IoU \\(\%)\end{tabular} & \begin{tabular}[c]{@{}l@{}}Acc \\(\%)\end{tabular} & \begin{tabular}[c]{@{}l@{}}Sens \\(\%)\end{tabular} & \begin{tabular}[c]{@{}l@{}}Spe \\(\%)~\end{tabular} & \begin{tabular}[c]{@{}l@{}}Prec \\(\%)\end{tabular} & \begin{tabular}[c]{@{}l@{}}DSC \\(\%)\end{tabular} & \begin{tabular}[c]{@{}l@{}}IoU \\(\%)\end{tabular} & \begin{tabular}[c]{@{}l@{}}Acc \\(\%)\end{tabular} & \begin{tabular}[c]{@{}l@{}}Sens \\(\%)\end{tabular} & \begin{tabular}[c]{@{}l@{}}Spe \\(\%)~\end{tabular} & \begin{tabular}[c]{@{}l@{}}Prec \\(\%)\end{tabular} &  &  &  \\ 
\hline
\multicolumn{16}{l}{CNN-based:} \\ 
\hline
Unet \cite{ronneberger2015u} & 88.42 & 81.22 & 95.32 & 90.85 & 96.85 & 89.67 & 89.26 & 81.42 & 92.86 & 95.62 & 93.01 & 85.43 & 31.04 & 54.74 & $256 \times 256$ \\
AttenUNet \cite{schlemper2019attention} & 88.8 & 81.77 & 95.45 & 90.58 & 96.77 & 90.49 & 89.3 & 81.76 & 92.75 & 95.72 & 92.58 & 85.76 & 34.88 & 66.63 & $256 \times 256$ \\
UNetPlusPlus \cite{zhou2018unet++} & 88.97 & 81.88 & 95.57 & 90.5 & 97.12 & 90.76 & 90.48 & 83.34 & 93.71 & 96.35 & \textbf{93.7} & \uline{86.77} & \textbf{9.16} & 34.90 & $256 \times 256$ \\
DualAttenUNet \cite{fu2019dual} & 88.9 & 81.98 & 95.53 & 91.27 & 96.74 & 90.32 & 90.39 & 83.19 & 93.72 & 96.91 & 92.88 & 86.12 & 19.57 & 28.88 & $256 \times 256$ \\ 
\hline
\multicolumn{16}{l}{Transformer-based:} \\ 
\hline
MISSFormer \cite{huang2022missformer} & 89.00 & 82.29 & 95.54 & 90.65 & 96.35 & 91.15 & \uline{91.45} & \uline{84.94} & \textbf{94.89} & 97.98 & 92.38 & 86.73 & 42.46 & 7.28 & $224 \times 224$ \\
SwinUNet \cite{cao2022swin} & 89.68 & 83.04 & 95.99 & 91.43 & 96.76 & 91.1 & 90.81 & 83.83 & 94.64 & \textbf{98.8} & 91.41 & 84.95 & 27.17 & 5.95 & $224 \times 224$ \\
H2Former \cite{he2023h2former} & 89.91 & 83.3 & \uline{96.07} & \uline{91.77} & 96.77 & 91.07 & 91.01 & 84.12 & 94.42 & 98.38 & 91.47 & 85.62 & 33.68 & 32.25 & $256 \times 256$ \\
HiFormer \cite{heidari2023hiformer} & 89.81 & 83.15 & \textbf{96.11} & 91.35 & 96.49 & 91.19 & 91.31 & 84.65 & 94.44 & \uline{98.44} & 91.47 & 86.11 & 37.46 & 17.51 & $224 \times 224$ \\
MulitTrans \cite{zhang2024multitrans} & \uline{90.03} & \uline{83.43} & \uline{96.07} & 90.95 & 96.9 & \uline{92.09} & 91.07 & 84.22 & 94.43 & 98.18 & 91.89 & 85.87 & 39.38 & 23.88 & $256 \times 256$ \\ 
\hline
\multicolumn{16}{l}{Mamba-based:} \\ 
\hline
Mamba\_UNet \cite{wang2024mamba} & 88.18 & 81.02 & 95.22 & 89.46 & 96.99 & 90.83 & 89.32 & 81.81 & 93.16 & 95.88 & 92.53 & 85.56 & 19.12 & 4.60 & $256 \times 256$ \\
VM\_UNet \cite{ruan2024vm} & 89.62 & 83.01 & 95.96 & 91.42 & 96.85 & 90.96 & 90.71 & 83.78 & 94.32 & 97.84 & 92.36 & 85.76 & 27.43 & \textbf{4.11} & $256 \times 256$ \\
VM\_UNet\_V2 \cite{zhang2024vm} & 89.51 & 82.7 & 95.9 & 90.47 & \textbf{97.2} & 91.44 & 91.27 & 84.55 & \uline{94.87} & 98.09 & 93.22 & 86.31 & 22.77 & \uline{4.40} & $256 \times 256$ \\
Swin\_UMamba \cite{liu2024swin} & 89.95 & 83.33 & 96.01 & \textbf{91.86} & 96.83 & 91.02 & 91.12 & 84.47 & 94.42 & 98.27 & 92.01 & 86.06 & 59.89 & 43.94 & $256 \times 256$ \\
Our & \textbf{90.12} & \textbf{83.51} & 96.04 & 90.75 & \uline{97.16} & \textbf{92.29} & \textbf{91.56} & \textbf{85.00} & 94.51 & 97.2 & \uline{93.61} & \textbf{87.57} & \uline{14.05} & 8.94 & $256 \times 256$ \\
\hline
\end{tabular}
  }
  \end{threeparttable}
  \end{table*}

\begin{figure}[!t]
  \centering
  \includegraphics[width=0.90\linewidth]{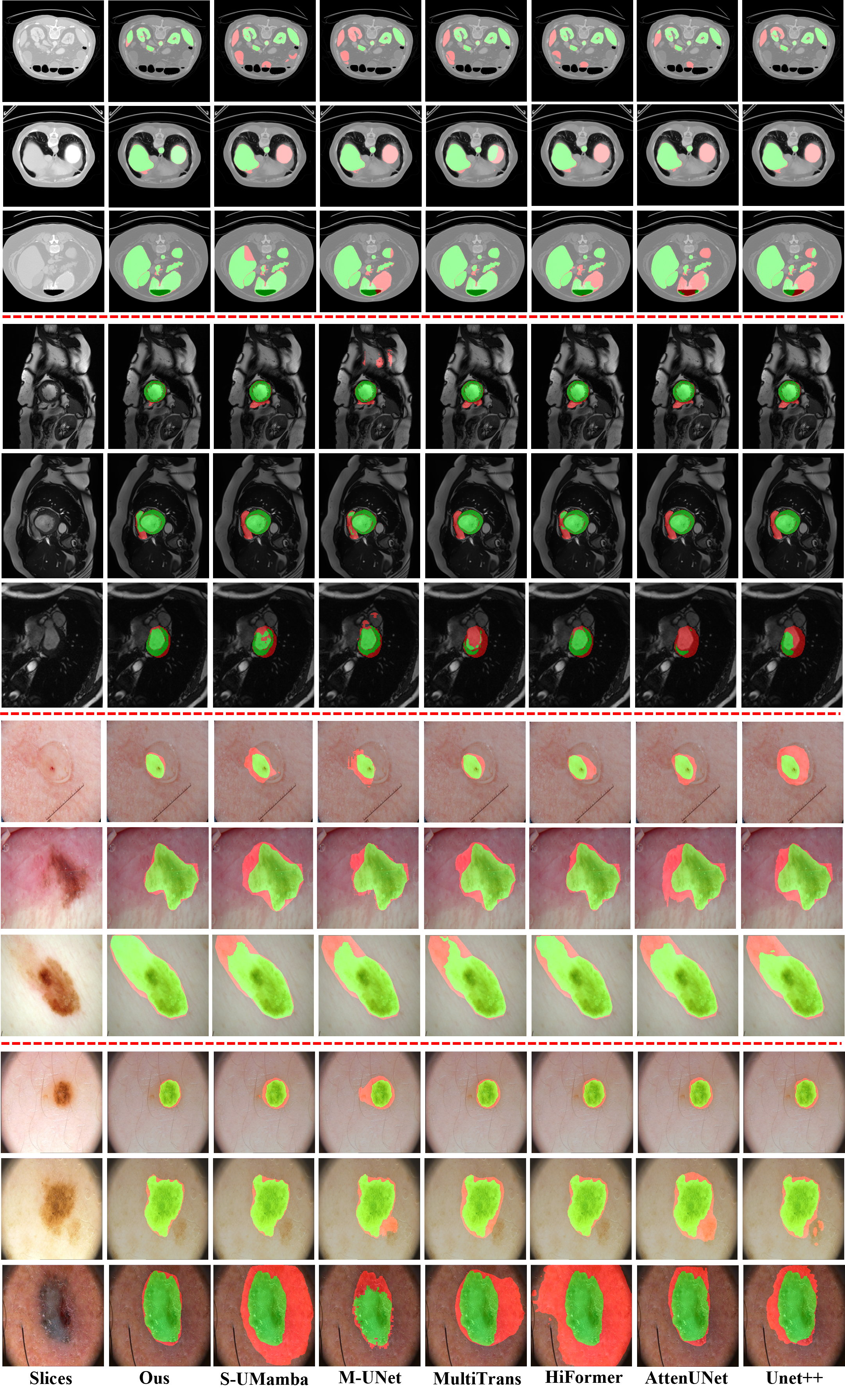}
  \caption{Visual comparisons with top-performing networks. From top to bottom, the results correspond to Synapse, ACDC, ISIC-2018, and PH2. S-UMamba denotes Swin-UMamba, and M-UNet refers to Mamba-UNet. Correct predictions are highlighted in green, and incorrect predictions are shown in red.} 
  \label{Visualization_Methods}
  \end{figure}


\section{Experiments}
\label{Experiments}

\subsection{Datasets}
\label{Datasets}

\subsubsection{Multi-organ Segmentation}  Synapse is a commonly used benchmark\footnote{https://www.synapse.org/\#!Synapse:syn3193805/wiki/217789}, including 30 axial abdominal clinical CT cases. As previously performed by other groups \cite{chen2021transunet}, 18 cases (2211 axial slices) are randomly selected for training and the remaining 12 cases for validation. Evaluation is conducted on eight abdominal organs, including the aorta (Ao), gallbladder (Ga), left kidney (Ki(L)), right kidney (Ki(R)), liver (Li), pancreas (Pa), spleen (Sp), and stomach (St). For training 2D networks, all sequences were resampled to an in-plane spacing of $0.7617 \text{ mm} \times 0.7617 \text{ mm}$, while the z-axis spacing remained unchanged. We clipped pixel intensities at the 0.5th and 99.5th percentiles within masks, corresponding to -956.0 and 277 HU, respectively, and normalized them using the training set mean (71.19 HU) and standard deviation (127.96). Finally, we applied random cropping of size $224 \times 224$ during training.  

\subsubsection{Cardiac Multi-structures Segmentation} The Automated Cardiac Diagnosis Challenge\footnote{https://www.creatis.insa-lyon.fr/Challenge/acdc/} (ACDC) dataset includes 200 cardiac MRI cases from 100 patients, including manual masks for the left ventricle (LV), the right ventricle (RV) and the myocardium (MYO). Following \cite{wang2022mixed, rahman2023medical}, the dataset was randomly partitioned at patient-level as follows: 70 patients for training, 10 for validation, and 20 for testing. For training 2D models, slices were extracted from the 3D volumes, resulting a total training set of 1,304 slices. During preprocessing, all sequences were resampled to a $1.5625 \text{ mm} \times 1.5625 \text{ mm}$ in-plane spacing, maintaining the original z-axis spacing. We applied Z-score normalization to each case independently and utilized random cropping to $192 \times 192$ for training.

\subsubsection{Skin Lesion Segmentation} ISIC 2018 \cite{codella2019skin} is a large public dataset containing 2,694 dermoscopy images collected from 432 patients. Following prior work \cite{ruan2022malunet, zhang2024vm}, we randomly divided them with a ratio of 7:3, resulting in 1,886 images for training and 808 for testing. Additionally, we used PH2 \cite{mendoncca2013ph}, a small external dataset containing only 200 dermoscopic images, to evaluate the generalization ability of the models. All images are resized to $256 \times 256$ pixels, normalized using the training set mean (155.25) and standard deviation (46.46), and subsequently scaled to the range [0,1]. The normalization statistics were computed over the resized images across the entire training set. 

\subsection{Evaluation Metrics}
\label{Evaluation Metrics}

Segmentation performance on the Synapse and ACDC datasets is evaluated using the Dice Similarity Coefficient (DSC), Intersection over Union (IoU), and 95\% Hausdorff Distance (HD). For skin lesion segmentation, in addition to DSC and IoU, we also report Accuracy (Acc), Sensitivity (Sens), Specificity (Spe), and Precision (Prec). Furthermore, we employ parameters (Params) and Floating Point Operations (FLOPs) to measure the model complexity and computational complexity, respectively. 

\subsection{Implementation Details}
\label{Implementation Details}

The experiments were conducted on PyTorch 2.0.1, with models trained on a single NVIDIA A800 (80GB) GPU. We optimized the model using the AdamW optimizer \cite{loshchilov2017decoupled} across all experiments. For the Synapse and ACDC datasets, we employed a batch size of 24, while a smaller batch size of 12 was used for ISIC2018. We set the weight decay to 1e-2 for Synapse and ISIC2018, and 5e-2 for ACDC. For learning rate scheduler, we used CosineAnnealingLR \cite{loshchilov2016sgdr} with a minimum learning rate of 1e-5. The maximum iterations and training epochs were set to 50 and 150 for the ACDC and ISIC-2018 datasets, and to 100 and 300 for the Synapse dataset. We applied random horizontal and vertical flip, random rotation from -180 to 180 degrees, and random scale from 0.7 to 1.4 for data augmentation. During inference, we adopt time-consuming evaluation tricks, including flipping and overlapped window sliding, to improve accuracy.  

To ensure a fair comparison across architectures, all networks were trained using the same protocol, with the learning rate found via grid search for each model. For our model, we found an optimal learning rate of 5e-4 for Synapse and ACDC, and 1.5e-4 for ISIC-2018.

\begin{table}[!t]\small
  \centering
  \caption{Ablation study of the Multi-level Feature Aggregation Module (MFAM) in the CNN part.}
  \label{Ablation study on MFAM.}
  \renewcommand\arraystretch{1.0}
  \begin{threeparttable}
  \setlength{\tabcolsep}{0.9mm}{
\begin{tabular}{lllllll} 
\hline
\begin{tabular}[c]{@{}l@{}}Top-down\\~path\end{tabular} & \begin{tabular}[c]{@{}l@{}}Bottom-up \\path\end{tabular} & \begin{tabular}[c]{@{}l@{}}DSC \\(\%)\end{tabular} & \begin{tabular}[c]{@{}l@{}}IoU \\(\%)\end{tabular} & \begin{tabular}[c]{@{}l@{}}HD95 \\(mm)\end{tabular} & \begin{tabular}[c]{@{}l@{}}Param \\(M)\end{tabular} & \begin{tabular}[c]{@{}l@{}}FLOPs \\(G)\end{tabular} \\ 
\hline
- & - & \begin{tabular}[c]{@{}l@{}}81.34 \\± 0.39\end{tabular} & \begin{tabular}[c]{@{}l@{}}72.77 \\± 0.17\end{tabular} & \begin{tabular}[c]{@{}l@{}}32.81 \\± 1.34\end{tabular} & 12.87 & 5.97 \\
- & \checkmark & \begin{tabular}[c]{@{}l@{}}81.62 \\± 0.85\end{tabular} & \begin{tabular}[c]{@{}l@{}}73.13 \\± 0.80\end{tabular} & \begin{tabular}[c]{@{}l@{}}31.05 \\± 1.63\end{tabular} & 13.32 & 6.12 \\
\checkmark & - & \begin{tabular}[c]{@{}l@{}}84.63 \\± 0.84\end{tabular} & \begin{tabular}[c]{@{}l@{}}76.59 \\± 1.09\end{tabular} & \begin{tabular}[c]{@{}l@{}}27.24 \\± 2.49\end{tabular} & 13.32 & 6.57 \\
\checkmark & \checkmark & \begin{tabular}[c]{@{}l@{}}85.04 \\± 0.50\end{tabular} & \begin{tabular}[c]{@{}l@{}}77.10 \\± 0.64\end{tabular} & \begin{tabular}[c]{@{}l@{}}28.05 \\± 3.49\end{tabular} & 14.05 & 6.87 \\
\hline
\end{tabular}
  }
  \end{threeparttable}
  \end{table}  

\begin{table}[!t]\small
  \centering
  \caption{Ablation study of the Recursive Alignment Module (RAM) for feature upsampling. Bilinear: bilinear interpolation. SA: Straightforward Alignment. Sup: Auxiliary supervision applied after the RAM.}
  \label{Ablation study on RAM.}
  \renewcommand\arraystretch{1.0}
  \begin{threeparttable}
  \setlength{\tabcolsep}{0.9mm}{
\begin{tabular}{llllll} 
\hline
Name & \begin{tabular}[c]{@{}l@{}}DSC \\(\%)\end{tabular} & \begin{tabular}[c]{@{}l@{}}IoU \\(\%)\end{tabular} & \begin{tabular}[c]{@{}l@{}}HD95 \\(mm)\end{tabular} & \begin{tabular}[c]{@{}l@{}}Param \\(M)\end{tabular} & \begin{tabular}[c]{@{}l@{}}FLOPs \\(G)\end{tabular} \\ 
\hline
Bilinear & \begin{tabular}[c]{@{}l@{}}83.85 \\± 0.81\end{tabular} & \begin{tabular}[c]{@{}l@{}}75.72 \\± 1.01\end{tabular} & \begin{tabular}[c]{@{}l@{}}30.25 \\± 3.46\end{tabular} & 14.00 & 6.82 \\
SFA & \begin{tabular}[c]{@{}l@{}}83.79 \\± 0.66\end{tabular} & \begin{tabular}[c]{@{}l@{}}75.62 \\± 0.82\end{tabular} & \begin{tabular}[c]{@{}l@{}}30.49 \\± 1.79\end{tabular} & 14.05 & 6.93 \\
w/o Sup & \begin{tabular}[c]{@{}l@{}}84.55 \\± 0.77\end{tabular} & \begin{tabular}[c]{@{}l@{}}76.47 \\± 0.79\end{tabular} & \begin{tabular}[c]{@{}l@{}}29.75 \\± 0.86\end{tabular} & 14.05 & 6.87 \\
RAM & \begin{tabular}[c]{@{}l@{}}85.04 \\± 0.50\end{tabular} & \begin{tabular}[c]{@{}l@{}}77.10 \\± 0.64\end{tabular} & \begin{tabular}[c]{@{}l@{}}28.05 \\± 3.49\end{tabular} & 14.05 & 6.87 \\
\hline
\end{tabular}
  }
  \end{threeparttable}
  \end{table}  

\begin{table}[!t]\small
  \centering
  \caption{Ablation study on the Pallel Mamba design: placing Mamba layers before or after the Recursive Alignment Module (RAM).}
  \label{Ablation study on Pallel Mamba.}
  \renewcommand\arraystretch{1.0}
  \begin{threeparttable}
  \setlength{\tabcolsep}{0.9mm}{
\begin{tabular}{lllllll} 
\hline
Position & Layers & \begin{tabular}[c]{@{}l@{}}DSC \\(\%)\end{tabular} & \begin{tabular}[c]{@{}l@{}}IoU \\(\%)\end{tabular} & \begin{tabular}[c]{@{}l@{}}HD95 \\(mm)\end{tabular} & \begin{tabular}[c]{@{}l@{}}Param \\(M)\end{tabular} & \begin{tabular}[c]{@{}l@{}}FLOPs \\(G)\end{tabular} \\ 
\hline
\begin{tabular}[c]{@{}l@{}}Before\\~RAM\end{tabular} & 2 & \begin{tabular}[c]{@{}l@{}}84.16 \\± 0.67\end{tabular} & \begin{tabular}[c]{@{}l@{}}75.94 \\± 0.74\end{tabular} & \begin{tabular}[c]{@{}l@{}}30.76 \\± 1.71\end{tabular} & 14.05 & 5.15 \\ 
\hline
\multirow{4}{*}{\begin{tabular}[c]{@{}l@{}}After \\RAM\end{tabular}} & 0 & \begin{tabular}[c]{@{}l@{}}83.44 \\± 0.81\end{tabular} & \begin{tabular}[c]{@{}l@{}}75.16 \\± 0.90\end{tabular} & \begin{tabular}[c]{@{}l@{}}29.68 \\± 2.58\end{tabular} & 12.7 & 4.31 \\
 & 1 & \begin{tabular}[c]{@{}l@{}}84.19 \\± 0.90\end{tabular} & \begin{tabular}[c]{@{}l@{}}76.04 \\± 1.07\end{tabular} & \begin{tabular}[c]{@{}l@{}}30.23 \\± 2.76\end{tabular} & 13.38 & 5.59 \\
 & 2 & \begin{tabular}[c]{@{}l@{}}85.04 \\± 0.50\end{tabular} & \begin{tabular}[c]{@{}l@{}}77.10 \\± 0.64\end{tabular} & \begin{tabular}[c]{@{}l@{}}28.05 \\± 3.49\end{tabular} & 14.05 & 6.87 \\
 & 3 & \begin{tabular}[c]{@{}l@{}}84.46 \\± 0.36\end{tabular} & \begin{tabular}[c]{@{}l@{}}76.37 \\± 0.50\end{tabular} & \begin{tabular}[c]{@{}l@{}}28.80 \\± 1.80\end{tabular} & 14.73 & 8.15 \\
\hline
\end{tabular}
  }
  \end{threeparttable}
  \end{table}  

\begin{table}[!t]\small
  \centering
  \caption{Ablation study of the Mamba-based Cross-scale Attention Module (MCAM). Sum denotes pixel-wise summation and SA represents spatial-attention for multi-scale fusion. CCI and CSI refer to the Cross-scale Channel and Cross-spatial Interaction operations, respectively.}
  \label{Ablation study on MPF.}
  \renewcommand\arraystretch{1.0}
  \begin{threeparttable}
  \setlength{\tabcolsep}{0.9mm}{
\begin{tabular}{llllllll} 
\hline
Name & CCI & CSI & \begin{tabular}[c]{@{}l@{}}DSC \\(\%)\end{tabular} & \begin{tabular}[c]{@{}l@{}}IoU \\(\%)\end{tabular} & \begin{tabular}[c]{@{}l@{}}HD95 \\(mm)\end{tabular} & \begin{tabular}[c]{@{}l@{}}Param \\(M)\end{tabular} & \begin{tabular}[c]{@{}l@{}}FLOPs \\(G)\end{tabular} \\ 
\hline
Sum & - & - & \begin{tabular}[c]{@{}l@{}}84.29 \\± 0.09\end{tabular} & \begin{tabular}[c]{@{}l@{}}76.11 \\± 0.11\end{tabular} & \begin{tabular}[c]{@{}l@{}}27.81 \\± 2.57\end{tabular} & 14.05 & 6.87 \\
SA & - & - & \begin{tabular}[c]{@{}l@{}}84.60 \\± 0.88\end{tabular} & \begin{tabular}[c]{@{}l@{}}76.46 \\± 1.19\end{tabular} & \begin{tabular}[c]{@{}l@{}}30.17 \\± 5.66\end{tabular} & 14.35 & 7.8 \\ 
\hline
\multirow{4}{*}{MCAM} & - & \checkmark & \begin{tabular}[c]{@{}l@{}}84.34 \\± 0.49\end{tabular} & \begin{tabular}[c]{@{}l@{}}76.13 \\± 0.51\end{tabular} & \begin{tabular}[c]{@{}l@{}}31.83\\~± 4.12\end{tabular} & 14.05 & 6.87 \\
 & \checkmark & - & \begin{tabular}[c]{@{}l@{}}84.55 \\± 0.38\end{tabular} & \begin{tabular}[c]{@{}l@{}}76.44 \\± 0.44\end{tabular} & \begin{tabular}[c]{@{}l@{}}30.86 \\± 0.54\end{tabular} & 14.05 & 6.87 \\
 & - & - & \begin{tabular}[c]{@{}l@{}}84.16 \\± 0.72\end{tabular} & \begin{tabular}[c]{@{}l@{}}75.84 \\± 0.87\end{tabular} & \begin{tabular}[c]{@{}l@{}}27.48 \\± 3.77\end{tabular} & 14.05 & 6.87 \\
 & \checkmark & \checkmark & \begin{tabular}[c]{@{}l@{}}85.04 \\± 0.50\end{tabular} & \begin{tabular}[c]{@{}l@{}}77.10 \\± 0.64\end{tabular} & \begin{tabular}[c]{@{}l@{}}28.05 \\± 3.49\end{tabular} & 14.05 & 6.87 \\
\hline
\end{tabular}
  }
  \end{threeparttable}
  \end{table}  

\begin{table}[!t]\small
  \centering
  \caption{Ablation for Multi-scale predictions. "-": The corresponding score map and weight map are discarded during the inference phase.}
  \label{Ablation for Multi-scale predictions.}
  \renewcommand\arraystretch{1.0}
  \begin{threeparttable}
  \setlength{\tabcolsep}{0.7mm}{
\begin{tabular}{lllllllll} 
\hline
Scale1 & Scale2 & Scale3 & Scale4 & \begin{tabular}[c]{@{}l@{}}DSC \\(\%)\end{tabular} & \begin{tabular}[c]{@{}l@{}}IoU \\(\%)\end{tabular} & \begin{tabular}[c]{@{}l@{}}HD95 \\(mm)\end{tabular} & \begin{tabular}[c]{@{}l@{}}Param \\(M)\end{tabular} & \begin{tabular}[c]{@{}l@{}}FLOPs \\(G)\end{tabular} \\ 
\hline
\checkmark & - & - & - & \begin{tabular}[c]{@{}l@{}}80.54 \\± 1.42\end{tabular} & \begin{tabular}[c]{@{}l@{}}70.80 \\± 1.90\end{tabular} & \begin{tabular}[c]{@{}l@{}}27.04 \\± 5.37\end{tabular} & \multirow{4}{*}{14.05} & \multirow{4}{*}{6.87} \\
- & \checkmark & - & - & \begin{tabular}[c]{@{}l@{}}83.49 \\± 1.10\end{tabular} & \begin{tabular}[c]{@{}l@{}}74.58 \\± 1.59\end{tabular} & \begin{tabular}[c]{@{}l@{}}36.57 \\± 2.60\end{tabular} &  &  \\
- & - & \checkmark & - & \begin{tabular}[c]{@{}l@{}}81.75 \\± 1.39\end{tabular} & \begin{tabular}[c]{@{}l@{}}72.43 \\± 1.56\end{tabular} & \begin{tabular}[c]{@{}l@{}}31.55 \\± 2.95\end{tabular} &  &  \\
- & - & - & \checkmark & \begin{tabular}[c]{@{}l@{}}82.51 \\± 1.17\end{tabular} & \begin{tabular}[c]{@{}l@{}}73.24 \\± 1.67\end{tabular} & \begin{tabular}[c]{@{}l@{}}32.94 \\± 6.13\end{tabular} &  &  \\ 
\hline
\checkmark & \checkmark & \checkmark & - & \begin{tabular}[c]{@{}l@{}}84.42 \\± 0.60\end{tabular} & \begin{tabular}[c]{@{}l@{}}76.21 \\± 0.80\end{tabular} & \begin{tabular}[c]{@{}l@{}}28.44 \\± 3.29\end{tabular} & \multirow{4}{*}{14.05} & \multirow{4}{*}{6.87} \\
\checkmark & \checkmark & - & \checkmark & \begin{tabular}[c]{@{}l@{}}84.97 \\± 0.43\end{tabular} & \begin{tabular}[c]{@{}l@{}}76.95 \\± 0.60\end{tabular} & \begin{tabular}[c]{@{}l@{}}27.14 \\± 5.23\end{tabular} &  &  \\
\checkmark & - & \checkmark & \checkmark & \begin{tabular}[c]{@{}l@{}}84.17 \\± 0.85\end{tabular} & \begin{tabular}[c]{@{}l@{}}75.99 \\± 1.10\end{tabular} & \begin{tabular}[c]{@{}l@{}}26.89 \\± 3.37\end{tabular} &  &  \\
- & \checkmark & \checkmark & \checkmark & \begin{tabular}[c]{@{}l@{}}84.97 \\± 0.48\end{tabular} & \begin{tabular}[c]{@{}l@{}}76.70 \\± 0.61\end{tabular} & \begin{tabular}[c]{@{}l@{}}32.48 \\± 5.10\end{tabular} &  &  \\ 
\hline
\checkmark & \checkmark & \checkmark & \checkmark & \begin{tabular}[c]{@{}l@{}}85.04 \\± 0.50\end{tabular} & \begin{tabular}[c]{@{}l@{}}77.10 \\± 0.64\end{tabular} & \begin{tabular}[c]{@{}l@{}}28.05 \\± 3.49\end{tabular} & 14.05 & 6.87 \\
\hline
\end{tabular}
  }
  \end{threeparttable}
  \end{table}  

\begin{figure}[!t]
  \centering
  \includegraphics[width=0.99\linewidth]{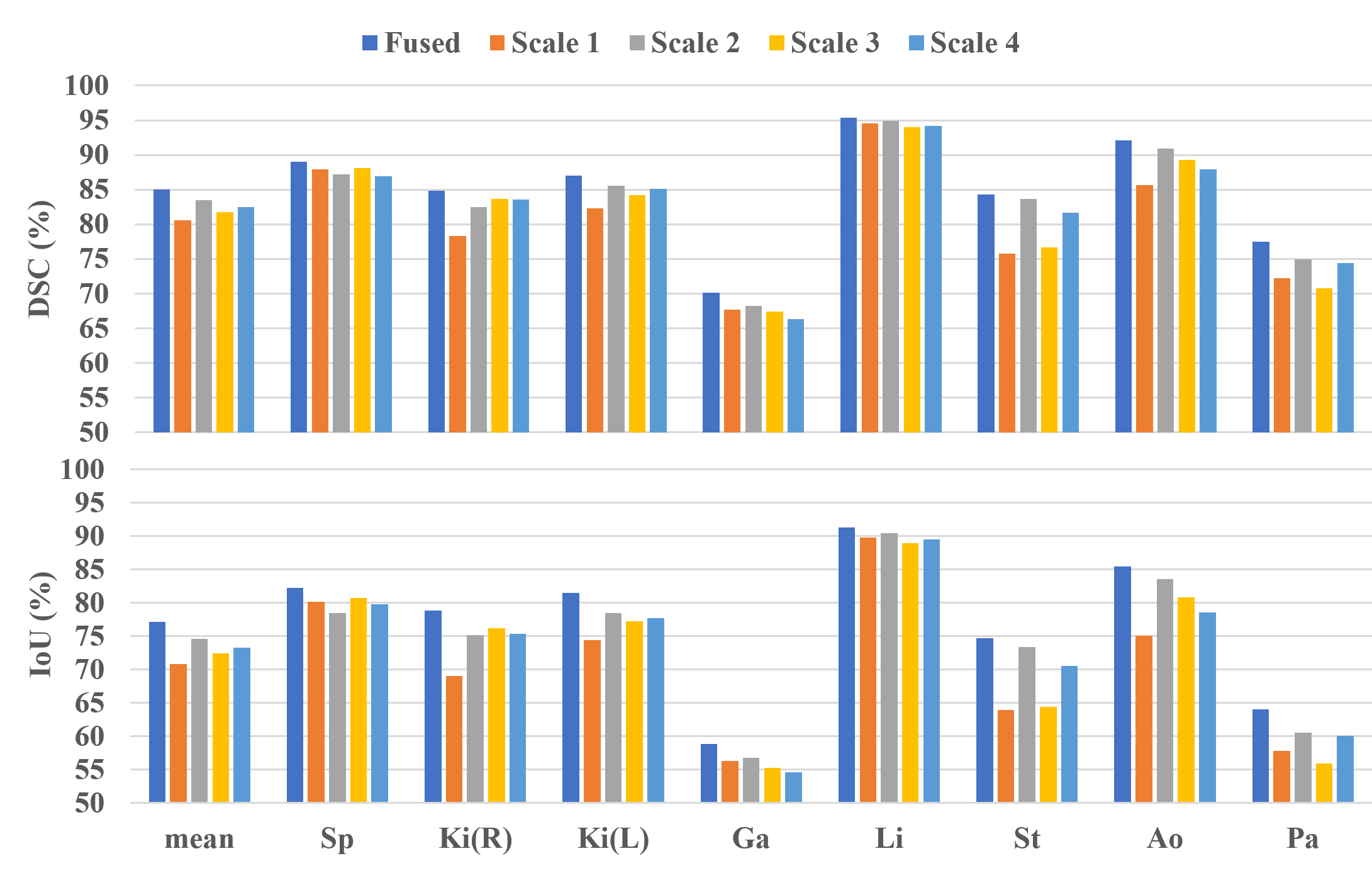}
  \caption{ Comparison of segmentation performance across abdominal organs, comparing individual scale predictions with the final integrated result.} 
  \label{Bat_chart}
  \end{figure}


\begin{figure}[!t]
  \centering
  \includegraphics[width=0.99\linewidth]{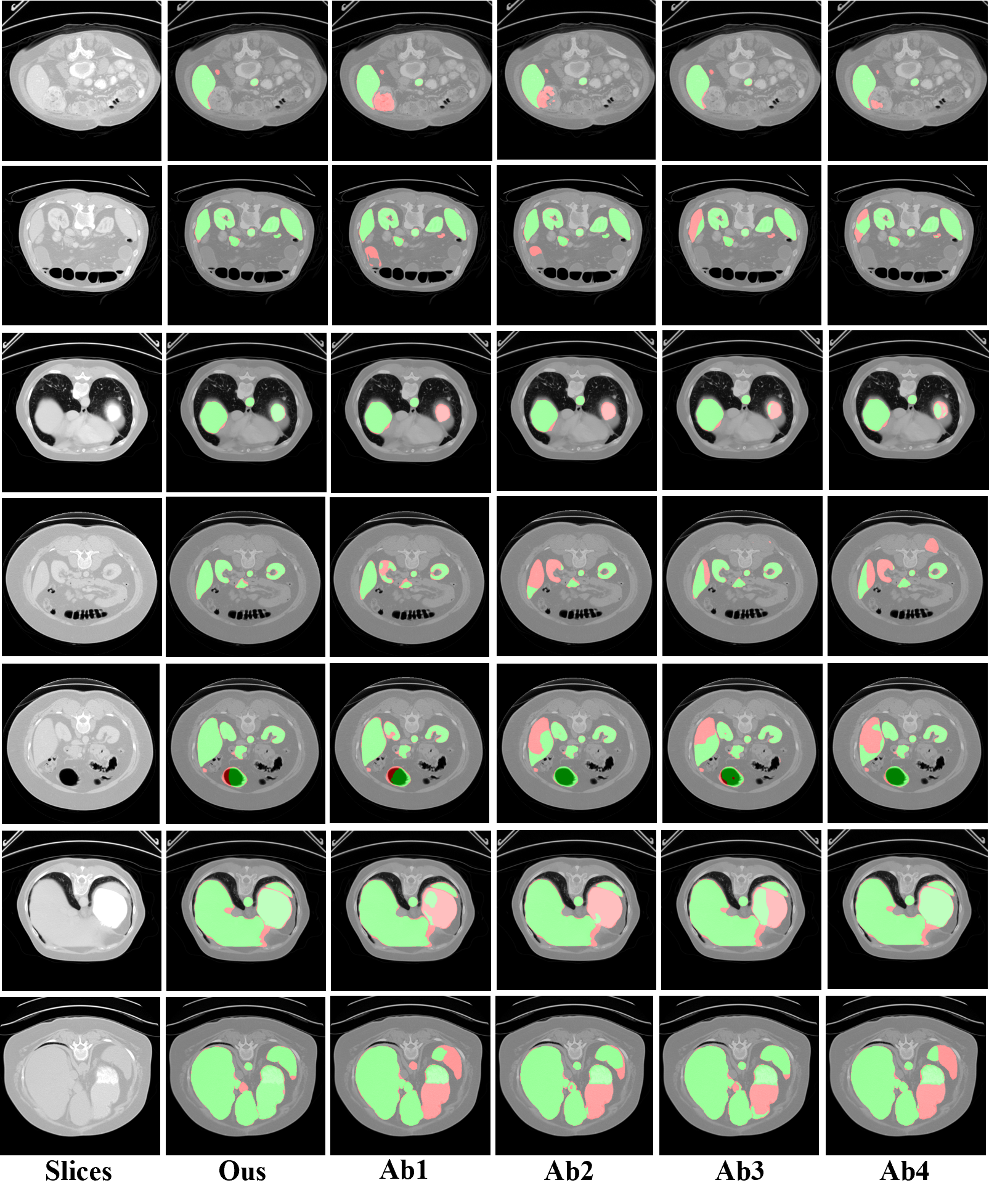}
  \caption{Qualitative results of ablation studies on the Synapse dataset. Ab1: w/o MFAM; Ab2: w/o RAM; Ab3: w/o Mamba layers; Ab4: w/o MCAM. Correct predictions are highlighted in green, and incorrect predictions are shown in red.} 
  \label{Visualization_Ablations}
  \end{figure}


\begin{figure}[!t]
  \centering
  \includegraphics[width=0.99\linewidth]{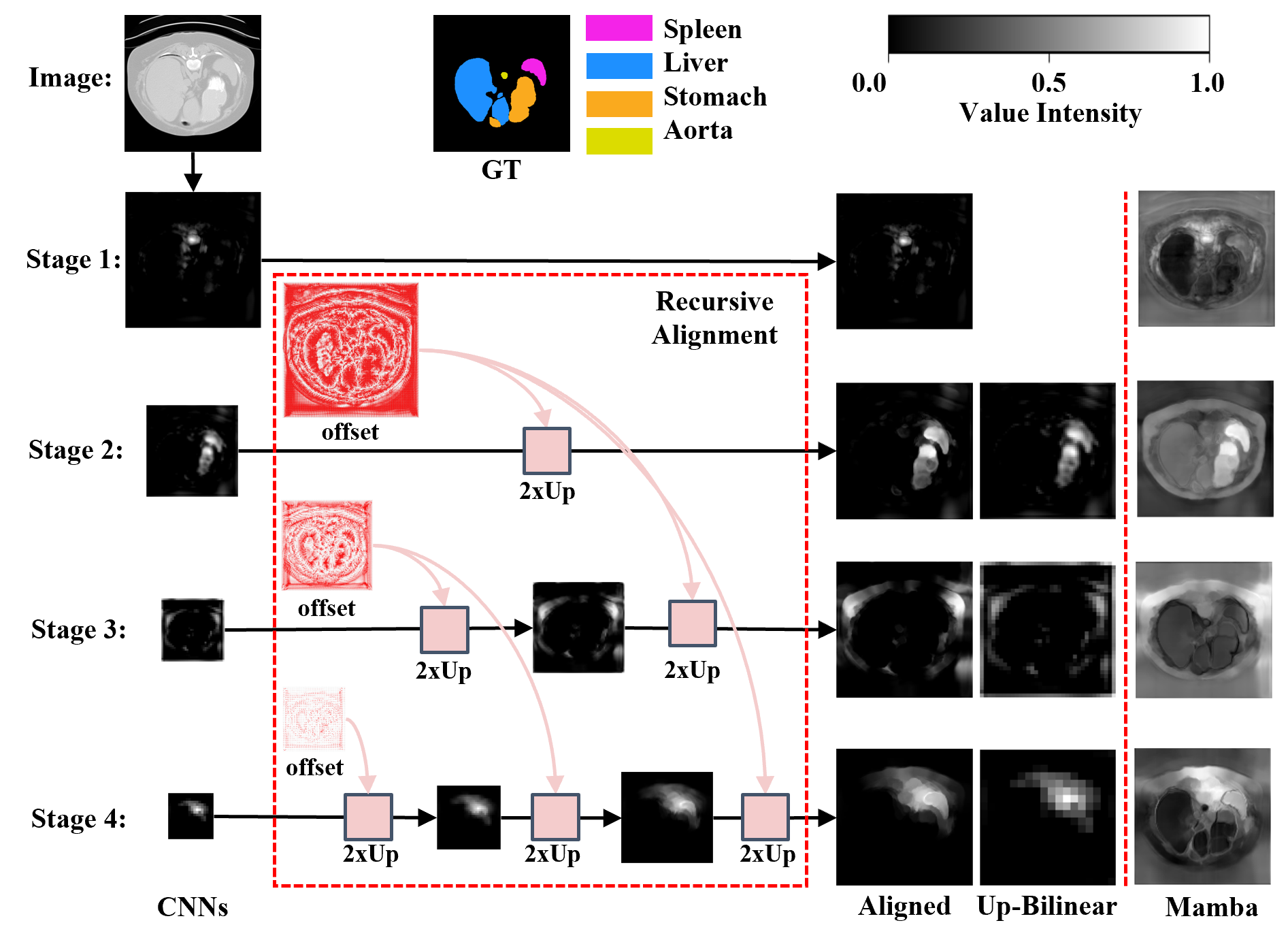}
  \caption{Visual analysis of recursive alignment and Mamba outputs. We show learned offsets per stage, compare aligned vs. bilinear upsampled features, and contrast Mamba-learned vs. CNN-learned features. To ensure comparability, feature maps are sampled from the same channel index of different stages and scales.} 
  \label{Visualization-Alignment}
  \end{figure}


\begin{figure}[!t]
  \centering
  \includegraphics[width=0.99\linewidth]{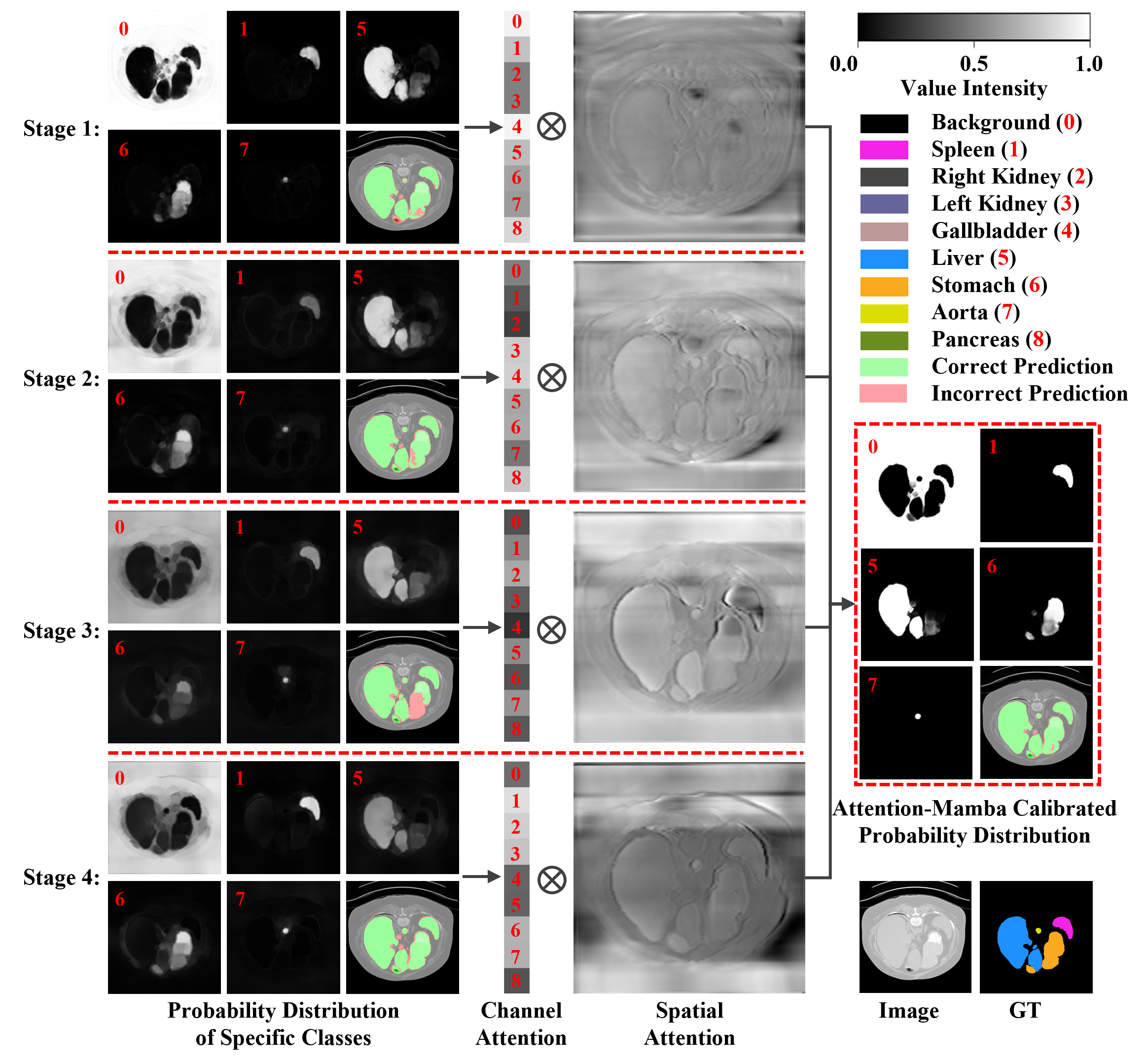}
  \caption{Visualization of the probability distributions for selected classes, along with channel and spatial attention intensities at each stage.} 
  \label{Visualization-Mamba-Attention}
  \end{figure}


\section{Results}

\subsection{Comparison with state-of-the-art methods}

Table. \ref{Table_Synapse} presents the results of our model for abdominal multi-organ segmentation. Compared with existing state-of-the-art approaches, the proposed method achieves improvements ranging from 0.84-14.17\% in DSC, 1.13-16.39\% in IoU, and 5.75-42.28 mm in HD95. Furthermore, it reaches the highest or second-highest DSC across all organs except the Aorta. 

Table. \ref{Table_ACDC} shows the results of our model on cardiac structure segmentation, compared to existing state-of-the-art methods. On the validation set, the proposed network improves DSC and IoU by 0.13-3.42\% and 0.28-5.62\%, respectively. On the test set, the gains are slightly higher, with DSC increasing by 0.18-3.8\% and IoU by 0.31-5.97\%. In addition, our model reaches the highest accuracy on all cardiac structures. 

Table. \ref{Table_ISIC-2018} shows results on skin lesion segmentation in both the internal and external test sets. Our proposed model achieves state-of-the-art performance across key metrics on both datasets. In particular, the network improves the DSC by 0.09-1.94\% and the IoU by 0.08-2.49\% on the ISIC-2018 test set. Furthermore, it demonstrates robust generalization on the external PH2 dataset, increasing the DSC and IoU by 0.11-2.30\% and 0.06-3.58\%, respectively. The model also yields the highest Precision across both datasets, meaning that it is able to reduce the number of false-positive predictions compared to existing methods. 

Regarding model size and computational complexity, our model, compared to the second-best method on each dataset, our model uses only a fraction of the parameters and FLOPs: 23.46\% and 20.46\% of Swin UMamba (Synapse), 35.68\% and 37.58\% of MultiTrans (ACDC and ISIC-2018), and 33.09\% and 94.37\% of MISSFormer (PH2), respectively. 

The qualitative results in Fig. \ref{Visualization_Methods} demonstrate the effectiveness of our network across diverse imaging modalities and varying object scales. Furthermore, the visualizations highlight the model's ability to reduce false-positive predictions.

\subsection{Ablation Studies}

\subsubsection{Quantitative Analysis}

Table. \ref{Ablation study on MFAM.} illustrates the performance gains achieved by the top-down and bottom-up paths. Integrating the top-down path enhances the DSC by 3.42\% and IoU by 3.97\%, requiring a minimal increase of 0.45 GFLOPs in complexity and 0.73 M parameters. Similarly, the bottom-up path improves DSC and IoU by 0.41\% and 0.51\%, respectively, while adding only 0.3 GFLOPs and 0.73 M parameters.

Table. \ref{Ablation study on RAM.} compares the performance of bilinear interpolation, straightforward alignment (SFA), and the proposed recursive alignment module (RAM) for upsampling low-resolution features. The RAM outperforms SFA, increasing the DSC and IoU by 1.25\% and 1.48\%, respectively. Compared to standard bilinear interpolation, the RAM yields improvements of 1.19\% in DSC and 1.38\% in IoU. Furthermore, introducing auxiliary supervision after the RAM provides an additional boost of 0.49\% (DSC) and 0.63\% (IoU), likely because it offers anatomical guidance for learning alignment offsets.

Table. \ref{Ablation study on Pallel Mamba.} evaluates the performance of the parallel Mamba design across various layer configurations and placements. The results indicate that incorporating two Mamba layers achieves the optimal balance between accuracy and model complexity. Specifically, this configuration outperforms the baseline (without Mamba layers) by 1.6\% in DSC and 1.94\% in IoU. In terms of placement, positioning the Mamba layers after the RAM (on upsampled features) proves more effective than placing them before the RAM (on low-resolution features). This choice yields an improvement of 0.88\% in DSC and 1.16\% in IoU, bring only a marginal increase in computational cost (1.72G FLOPs).

The ablation results for the multi-scale prediction fusion design are summarized in Table. \ref{Ablation study on MPF.}. Spatial Attention (SA) utilizes attention heads composed of several convolutional layers to independently generate attention scores for each stage's prediction. Compared to pixel-wise summation fusion (Sum), our proposed MCAM improves the DSC and IoU by 0.75\% and 0.99\%, respectively, with negligible impact on model size or computational overhead. When compared to SA, our method achieves gains of 0.44\% in DSC and 0.64\% in IoU, while simultaneously reducing parameters by 0.3M and computational complexity by 0.93 GFLOPs. Regarding the specific components of MCAM, the inclusion of CCI yields a 0.7\% increase in DSC and a 0.97\% increase in IoU. Similarly, CSI improves DSC by 0.49\% and IoU by 0.66\%; notably, both modules maintain the model's efficiency without significant increases in size or complexity.

Table. \ref{Ablation for Multi-scale predictions.} highlights the relative importance of each scale to the overall prediction. The predictions from scale 2 (DSC: 83.49\%, IoU: 74.58\%) and scale 4 (DSC: 82.51\%, IoU: 73.24\%) are notably more accurate than those from scale 1 (DSC: 80.54\%, IoU: 70.80\%) and scale 3 (DSC: 81.75\%, IoU: 72.43\%). This observation is consistent with the results of removing one of the four scales: removing scale 2 results in a 0.84\% drop in DSC and a 1.11\% drop in IoU, while removing scale 4 leads to decreases of 0.62\% in DSC and 0.89\% in IoU. In contrast, removing scale 1 causes only a 0.07\% drop in DSC and 0.40\% in IoU, and removing scale 3 yields decreases of 0.07\% in DSC and 0.15\% in IoU.

Fig. \ref{Bat_chart} shows that the segmentation accuracy of the final prediction consistently exceeds that of any single-scale prediction across all abdominal organs, demonstrating the effectiveness of the MCAM for adaptive multi-scale predictions fusion.

\subsubsection{Qualitative Analysis}

Visualizations in Fig. \ref{Visualization_Ablations} demonstrate that the integration of MFAM, RAM, Mamba layers, and MCAM effectively reduces incorrect predictions across objects with various scales.

In Fig. \ref{Visualization-Alignment}, the learned offsets generate deformable fields that closely follow the anatomical contours of the organs; consequently, the aligned features exhibit noticeably clearer boundaries shapes than the blurred results of those upsampled via bilinear interpolation. Additionally, while the CNN features remain focused within localized regions, the Mamba layers effectively expand the activation areas, providing the output features with a global receptive field.

Fig. \ref{Visualization-Mamba-Attention} illustrates the class probability distributions and predictions across each stage. The proposed attention-Mamba mechanism effectively calibrates these distributions, thereby reducing prediction errors. Furthermore, the channel and spatial attention maps demonstrate how the mechanism selectively focuses on or suppresses specific channels (classes) and regions of each stage.

\section{Discussion}
\label{Discussion}

In this work, we proposed an innovative framework able to consistently outperform representative CNN-, Transformer-, and Mamba-based methods across four datasets. Moreover, this trend is observed across different imaging modalities, including MRI, CT, and dermoscopy, suggesting robustness to modality- and task-specific variations, indicating good generalizability of the algorithm. Although the performance gains over the second-best methods are modest, the proposed network achieves a favorable trade-off between accuracy, model size, and computational efficiency. This makes it particularly suitable for resource-constrained clinical settings. Qualitative results in Fig. \ref{Visualization_Methods} further illustrate its effectiveness in handling multi-scale targets and reducing false positives.

In addition, ablation studies show that both the top-down and bottom-up paths improve segmentation performance with only a marginal increase in model size and complexity. This demonstrates the importance of providing multi-level features for the prediction of each scale. Regarding feature upsampling, Table. \ref{Ablation study on RAM.} shows that our proposed RAM is more effective than straightforward alignment (SFA) and bilinear interpolation. As seen in the visualizations (Fig. \ref{Visualization-Alignment}), the RAM aligns low-resolution features by following the anatomical structure of organs, and the comparison with bilinear upsampled features confirms the necessity of aligning features before multi-scale fusion. The positioning of the Mamba branches is also important. Placing Mamba layers after the RAM (on upsampled features) is more effective than placing them before (on low-resolution features). This is likely because Mamba is well-suited for processing long sequences but has limitations with short sequences, as noted by \cite{yu2025mambaout}. Furthermore, Table. \ref{Ablation study on MPF.} and Fig. \ref{Visualization-Mamba-Attention} demonstrate that channel and spatial interactions between multi-scale features are important for generating accurate attention maps, and the MCAM can effectively reduce incorrect predictions by highlighting or suppressing certain channels and spatial regions at each scale.

Multi-scale prediction is the core design of our work, distinct from the widely adopted U-shaped architecture. Detailed ablation studies in Table. \ref{Ablation for Multi-scale predictions.} indicate the importance of each scale, specifically showing that scales 2 and 4 contribute more than scales 1 and 3. As Fig. \ref{Bat_chart} illustrates, the prediction of different scales is better at different organs and the segmentation accuracy of the fused prediction consistently exceeds that of any single-scale prediction across all abdominal organs, further demonstrating the rationality of this multi-scale predication design.

The study has several limitations. We currently focused on a 2D version of the proposed architecture due to limited computational resources. In future work, we will extend the model to a 3D version and conduct comparisons with existing 3D models. Furthermore, the model design still leaves room for improvement. To enable convenient feature fusion in the MFAM, all encoder features are adjusted to the same number of channels using a 1 $\times$ 1 convolutional layer. Additionally, each Mamba branch contains identical layers across different scales. The proposed architecture has the potential to achieve a better balance between performance and efficiency by assigning specific channel dimensions and numbers of Mamba layers according to the characteristics of different scales.

In conclusion, within this work, we demonstrated that combining multi-scale parallel inference with Mamba-based modeling is an effective strategy for improving segmentation performance while maintaining computational efficiency.



\section*{Statements of Ethical Approval}
\label{Statements of ethical approval}
We used three published datasets, so ethical approval was not required.

\section*{Declaration of Competing Interest}
\label{Declaration of Competing Interest}
The authors declare that they have no known competing financial interests or personal relationships that could have appeared to influence the work reported in this paper.


\bibliographystyle{elsarticle-num} 
\bibliography{References_MFARANetV2.bib}


\end{document}